\documentclass{article}
\usepackage{arxiv}

\usepackage[utf8]{inputenc} 
\usepackage[T1]{fontenc}    
\usepackage{url}            
\usepackage{booktabs}       
\usepackage{amsfonts}       
\usepackage{nicefrac}       
\usepackage{microtype}      
\usepackage{lipsum}
\usepackage{array, makecell}
\usepackage{graphics}
\usepackage{graphicx}
\usepackage{tikz}
\usetikzlibrary{tikzmark}
\usepackage{comment}
\usepackage{url}
\usepackage{subcaption}
\usepackage{pgfplots}
\pgfplotsset{compat=newest}
\usepackage{mathtools}

\usepackage{multirow}
\usepackage[colorlinks=true,linkcolor=blue,urlcolor=blue,citecolor=blue]{hyperref}
\usepackage{algorithmic}
\usepackage[ruled,vlined,linesnumbered]{algorithm2e}
\SetKwComment{Comment}{$\triangleright$\ }{}

\title{An Ensemble-based Multi-Criteria Decision Making Method for COVID-19 Cough Classification} 
\author{
 Nihad Karim Chowdhury\thanks{Corresponding Author}\\
  Department of Computer Science and Engineering\\
  University of Chittagong\\
  Bangladesh\\ 
  \texttt{nihad@cu.ac.bd}\\
  \And
 Muhammad Ashad Kabir\\
 School of Computing and Mathematics\\
 Charles Sturt University, NSW\\
 Australia\\ 
 \texttt{akabir@csu.edu.au} \\
 \And
 Md. Muhtadir Rahman\\
 Department of Computer Science and Engineering\\University of Chittagong\\
 Bangladesh\\ 
 \texttt{muhtadir.cse@std.cu.ac.bd} \\
  
}

\begin{document}
\maketitle

\begin{abstract}

%The deployment of screening tools that can quickly provide test results, can be used to combat the second wave of the COVID-19 pandemic, as the infection spreads rapidly due to its contagious nature.
%With the current functional world suddenly halting, it is necessary to provide results within a fast turnaround time and rapidly scale up diagnostic tests.
%To this end, we have explored the sound of coughing as a diagnostic test to identify potential COVID-19 infection.
The objectives of this research are analysing the performance of the state-of-the-art machine learning techniques for classifying COVID-19 from cough sound and identifying the model(s) that consistently perform well across different cough datasets. Different performance evaluation metrics (such as precision, sensitivity, specificity, AUC, accuracy, etc.) make it difficult to select the best performance model. To address this issue,
in this paper, we propose an ensemble-based multi-criteria decision making (MCDM) method for selecting top performance machine learning technique(s) for COVID-19 cough classification.
We use four cough datasets, namely Cambridge, Coswara, Virufy, and NoCoCoDa to verify the proposed method.
At first, our proposed method uses the audio features of cough samples and then applies machine learning (ML) techniques to classify them as COVID-19 or non-COVID-19.
Then, we consider a multi-criteria decision-making (MCDM) method that combines ensemble technologies (i.e., soft and hard) to select the best model.
In MCDM, we use technique for order preference by similarity to ideal solution (TOPSIS) for ranking purposes, while entropy is applied to calculate evaluation criteria weights.
In addition, we apply the feature reduction process through recursive feature elimination with cross-validation under different estimators. 
The results of our empirical evaluations show that the proposed method outperforms the state-of-the-art models.
We see that when the proposed method is used for analysis using the Extra-Trees classifier, it has achieved promising results (AUC: $0.95$, Precision: $1$, Recall: $0.97$).
%We hope that the proposed method will facilitate the development of a fully automated and effective COVID-19 cough detection system.

\end{abstract}

% keywords can be removed

\section{Introduction}
\label{sec:Introduction}

%As of $31$ May $2021$, Corona Virus Disease (COVID-19) infection has caused $3,540,437$ deaths, while the number of confirmed cases is $170,051,718$, and has interrupted lives in $223$ countries throughout the world~\cite{26}.
With the outbreak of the second wave of COVID-19 pandemic, losses in all aspects of human life are increasing every day.
As we have observed, a destructive second wave is destroying some country's health care systems and crematoriums.
To limit the spread of the virus, regional regular testing and contact tracing can substitute regional restraints~\cite{27}, and the “Trace, Test and Treat” policy had flattened the pandemic trajectory (for instance, in Singapore, South Korea and China) in its initial stages~\cite{18}.
Therefore, in order to reduce the infection rate and limit the impact of medical resources, fast and economical COVID-19 infection detection methods are indispensable.  
Infected countries have implemented many strategies to limit the spread of this virus.
Such strategies include, encouraging people to maintain social distancing and personal hygiene, enhancing infection screening systems through multi-functional testing, and pursuing mass vaccination to reduce the pandemic ahead of time, etc.
Developing or underdeveloped countries are still striving to improve their detection capabilities because current methods of detecting COVID-19 (such as reverse transcription-polymerase chain reaction (RT-PCR)) require the use of expensive kits for on-site testing, and these kits are not always easy to obtain.
Hence, low-cost, distributable, and reliable pre-screening tests are essential for identifying and diagnosing COVID-19 and limiting local outbreaks of COVID-19 infection.

Besides, RT-PCR standard diagnostic scheme, several artificial intelligence (AI)-based methods have recently been proposed that use chest X-rays (\cite{29,28,56}) and CT scans (\cite{30,31}) to distinguish COVID-19 from other bacterial/viral infections.
At the same time, to use RT-PCR, CT scan and X-ray for diagnosis, it is essential to go to the testing center, and well-equipped clinical facilities.
Since the above-mentioned test protocol involves multiple people at close range, there is a high risk of spreading infection to a greater extent due to the infectivity of COVID-19.
To limit the exponential growth of the number of COVID-19 cases, one solution is to design a model that can perform biological tests without intervening many people.
Therefore, many AI-based applications that use audio with less human contact have been used for testing and early detection of respiratory diseases. 
As we all know, cough is a distinctive symptom of many respiratory diseases, and cough symptoms have been used to detect different types of respiratory diseases such as pulmonary edema, tuberculosis, pneumonia, whooping cough, and asthma through AI-based models~\cite{59,60,61,62}.
It is prevalent that COVID-19 infects the respiratory system, affecting the sound of someone's coughing, breathing, and voice tone.
Recently, a number of research have proposed audio-based AI models~\cite{18,1,2,4,21,25,57,58} for detecting the infection status of COVID-19.

%As we have observed in some previous studies (\cite{18,1,2,4,21,25,57,58}), COVID-19 or non-COVID-19 coughs from various respiratory syndromes have unique underlying characteristics, which are obtained through the conversion of cough sounds.
%To limit the exponential growth of the number of COVID-19 cases, one solution is to design a model that can perform biological tests through mobile or web applications without intervening many people.
%In addition, isolate those who self-test positive so that infection can be restricted. 

In this paper, we propose a machine learning (ML)-based COVID-19 detection architecture using audio recordings, in particular, cough sound.
Our work includes the use of crowdsourcing data from the University of Cambridge~\cite{1}, which contains two categories, namely asymptomatic and symptomatic, to explore the use of human coughing as a unique marker of COVID-19.
Subsequently, we validate the proposed method using other datasets, such as Coswara~\cite{2}, Virufy~\cite{57}, and Virufy integrated with NoCoCoDa~\cite{17}.
The key idea of our work is to generate audio features, such as Mel-Frequency Cepstral Coefficient (MFCC), Chromagram, Mel-Scaled Spectrogram, Spectral Contrast and Tonal Centroid, before inputting the data to a classifier while maintaining a high level of detection performance acceptable to COVID-19 cases.
We then use some popular ML-based classification techniques for binary classification (i.e., categorizing between COVID-19 and non-COVID-19).
After that, we consider using a multi-criteria decision (MCDM)~\cite{52} method to evaluate the results of each classification technique and consider three separate training strategies with different frameworks and hyper-parameter choices (see in Section~\ref{sec:Training strategies and hyper-parameters optimization}).
Entropy is considered for selecting weights of different evaluation criteria, and then the generated weights are assigned to the weights used for Technique for Order Preference by Similarity to Ideal Solution (TOPSIS)~\cite{51}, which are used for the ranking of the models in the MCDM method.
The MCDM outputs from each training strategy are
aggregated through soft and hard ensemble to make the best decision for choosing the best model.

Indeed, model comparisons that only consider one or a few evaluation criteria (i.e., accuracy, precision, etc.) cannot reflect the actual model performance when the dataset is imbalanced.
Therefore, we consider MCDM that deals with various evaluation criteria, and to select the best model.
Moreover, MCDM has proven its effectiveness in some aspects of the COVID-19 management system~\cite{54,55}.
Also, we have integrated ensemble methods in MCDM frameworks, thereby reducing the decision bias in choosing the best model. 
To the best of our knowledge, this is the first attempt to explore ensemble-based MCDM in detecting COVID-19 from cough sound.
Furthermore, to support the development of the proposed architecture, we perform an extensive experiment through Recursive Feature Elimination with Cross-Validation (RFECV) to rank audio features.
By using the top-ranked features, we have increased the AUC score of the asymptomatic category by $3\%$ and the AUC score of the symptomatic category by $12\%$ compared to the baseline AUC score without feature selection.
%Finally, we use two independent datasets~\cite{1,2} for the validation process so that we can generalize the results of the study.
%In this kind of cross-institutional analysis, we use one dataset to build the proposed model while fixing other dataset for testing.
The research results show that our proposed architecture can effectively detect COVID-19 cough.
In addition, the results of the ensemble-based MCDM results of different ML models can help medical practitioners to choose the best performing model under different experimental settings. 
The main contributions of this paper are summarized as follows:
\begin{itemize}
  \item We propose an ensemble-based MCDM method for detecting COVID-19 from cough sound data.
  \item We propose three testing strategies with different frameworks and hyper-parameter optimization to analyze the existing baseline ML models' detection performance for identifying the best model.
  \item We apply feature selection methods to identify the most important features, thereby significantly improving prediction performance.
  \item We consider four independent cough datasets for validation to confirm the effectiveness of our proposed method. 
  \item We conduct an empirical evaluation of the model and compare it with the state-of-the-art models to evaluate the effectiveness of the proposed method in distinguishing COVID-19 from non-COVID-19.
  
  %\item Considering the method of selecting the best COVID-19 diagnostic model, we integrate entropy and TOPSIS into the proposed architecture as a measure of MCDM.
  %\item We propose an ensemble strategy that takes into account the multiple results of MCDM from multiple testing strategies to help doctors determine the best model so that the system can be evaluated reliably and impartially.
  %\item We analyze the experimental evaluation of asymptomatic and symptomatic categories through hybrid and cross-training and testing approaches to confirm the validity of our proposed architecture.  
  
\end{itemize}

The rest of this article is organized as follows.
Section~\ref{sec:Related Works} presents related work, and Section~\ref{sec:Methodology} describes methodology and explains our proposed  method. Section~\ref{sec:resultanalysis} report our experimental results. Finally, Section~\ref{sec:conclusion} summarizes the paper with future work.

\section{Related Works}
\label{sec:Related Works}

%Due to the development of digital technology, audio signals generated by the human body have been collected through digital technology, and the collected data are automatically analyzed.
%Sorting only from audio can be simple to operate and easy to distribute.
Early research~\cite{9,10} findings indicate that coughs originating from specific infections or diseases have sufficient distinguishing characteristics that ML-based models can use for classification.
Furthermore, several ML-based methods~\cite{12,13,14,15} have shown significantly superior performance in using sound to diagnose various respiratory diseases in automatic audio interpretation.

Nowadays, many researchers have begun to explore the respiratory sounds (i.e., cough, breath and voice)  of patients who have tested positive for COVID-19 and try to distinguish them from healthy people's sounds.
In the first step, it needs to create a valid audio benchmark dataset to effectively diagnose COVID-19. 
Many researchers have made significant efforts to create dataset such as Cambridge University sound data~\cite{1}, Coswara~\cite{2}, Cough against COVID~\cite{4}, COVID-19 cough dataset~\cite{21}, AI4COVID~\cite{18}, COUGHVID~\cite{25}, Virufy~\footnote{\label{note1}\url{https://github.com/virufy/virufy-data}}~\cite{57}, Novel Coronavirus Cough Database (NoCoCoDa)~\cite{17}, Breathe for Science~\footnote{\url{https://www.breatheforscience.com}}, and SARS COVID-19 in South Africa (Sarcos)~\footnote{\url{https://coughtest.online}}.
With their release, several pieces of research have been conducted that focus on the ML-based COVID-19 detection model from audio samples.
We divide the literature review of ML-based COVID-19 detection methods based on audio samples into four groups: 1) speech and voice, 2) cough, breadth and voice, 3) cough and breadth, and 4) cough only.
In the following, we review the most relevant research work. 

%A few studies~\cite{20,64} on the AI-based COVID-19 detection model to this date have distinguished COVID-19 positive from COVID-19 negative using only speech and voice samples.
Some studies~\cite{64,68} used only speech and voice sounds for classifying COVID-19.
%In~\cite{20}, Jing et al. analyzed speech recordings of COVID-19 patients to classify the health status of patients from four perspectives, such as the severity of illness, sleep quality, fatigue, and anxiety.
%To do this, they used two established acoustic feature sets and support vector machines.
A few other studies~\cite{2,47} have attempted to explore cough, breath, and voice samples toward the analysis of COVID-19 detection.
%In addition, the Coswara~\footnote{\url{https://github.com/iiscleap/Coswara-Data}} dataset was developed by Sharma et al.~\cite{2} to build a diagnostic tool to detect COVID-19 using breath sounds (deep and shallow), cough sounds (heavy and shallow), sustained phonation of vowel sounds (/a/ as in made, /i/,/o/), and count from $1$ to $20$ (fast and normal).
Some studies~\cite{1,22,23,38} use cough and breath samples as diagnostic symptoms for COVID-19 testing.
In~\cite{1}, Brown et al. proposed a binary predictive model in which they used cough and breath to distinguish the sound of COVID-19 from asthma or healthy.
They extracted audio features and combined them with the output of a pre-trained audio neural network. Their model achieved a receiver operating characteristic - area under curve (ROC-AUC) of over $0.80$ in all tasks designed during the experiment.
In~\cite{22}, the raw breath and cough audio and spectrogram were used to identify whether the patient was infected with COVID-19 through the ensemble of neural networks.
Here, the combination of Bayesian optimization and hyperband was considered for automatic hyper-parameter selection, which achieved an unweighted average recall rate (UAR) of $0.74$ or an AUC of $0.80$.
Harry et al.~\cite{23} proposed a novel modeling approach that utilizes a custom deep neural network based on ResNet~\cite{24} to diagnose COVID-19 from mutual breathing and cough representation, with an AUC of $0.846$.
QUCoughScope~\cite{38} is a mobile application that uses Cambridge University dataset to automatically detect asymptomatic COVID-19 patients using the cough and breathing sounds.
%In this work, Chowdhury et al. implemented a deep learning pipeline on the back end, which can immediately present the results to a mobile phone application to highlight the respiratory tract, such as a COVID-19 infection or a healthy respiratory tract.

Many studies~\cite{4,37,21,18,41,53,57,63,67} considered the analysis of cough audio signals as a workable course of action for an initial COVID-19 diagnosis.
In~\cite{4}, cough sounds were analyzed through an AI-based model, and the proposed model showed a statistically significant signal, indicating the status of COVID-19.
%Here, the author used microbiologically confirmed COVID-19 coughs from $3,621$ individuals and obtained an AUC score of $0.72$ using the CNN architecture (i.e., ResNet18).
Here, the authors used microbiologically confirmed COVID-19 coughs and obtained an AUC score of $0.72$ using the CNN architecture ResNet18. 
Using cough sounds, Ankit et al.~\cite{37} proposed an AI framework for diagnosing COVID-19 with interpretable features. 
The proposed framework combined cough sound characteristics with patient symptoms during empirical evaluation, and included four cough categories, such as COVID-19, asthma, bronchitis and healthy.
In~\cite{21}, the AI speech processing framework for COVID-19 is pre-screened from cough records using the speech biomarker feature extractor.
In this method, cough records are converted by MFCC and put into a CNN-based architecture, which consists of a Poisson biomarker layer and three pre-trained ResNet50’s~\cite{24} in parallel.
Imran et al.~\cite{18} proposed a model called AI4COVID, which can distinguish the pathomorphological changes caused by COVID-19 infection in the respiratory system and compare it with other respiratory infections (such as pertussis and bronchitis) and normal respiratory tract.
Also, the authors developed a tri-pronged mediator-centered AI engine to reduce the misdiagnosis risk for the cough-based diagnosis of COVID-19.
Madhurananda et al.~\cite{41} used two datasets, Coswara and Sarcos, to diagnose COVID-19 from cough samples.
The authors explored seven ML-based approaches, and from empirical evaluation, it had been shown that ResNet50 and LSTM got higher AUC scores compared with the other ML methods.
Javier et al.~\cite{53} proposed a COVID-19 cough detection algorithm based on empirical mode decomposition (EMD), and then introduced the acoustic sonography tensor and a deep artificial neural network classifier with convolutional layers for subsequent classification.
%It validated the detection performance of the proposed method with $8,380$ clinically validated samples, i.e., $2,339$ COVID-19 positive and $6,041$ COVID-19 negative, with the laboratory molecular-test under quantitative RT-PCR (qRT-PCR).
Another work~\cite{63} developed a classifier for the COVID-19 pre-screening model from two publicly available crowd-sourced cough sound samples, in which they divided the cough sound samples into non-overlapping coughs, and it extracted six cough features from each.
The authors conducted a lot of experiments on shallow ML, convolutional neural networks (CNN) and pre-trained CNN models, and reported that an ensemble of CNN can achieve better accuracy.  
%The author reports that better accuracy is obtained using the ensemble method, where the kappa statistic ($>= 0.2$) is used to select candidate classifiers.

There are some limitations accompanied by the previous studies.
Previous studies have used a number of evaluation criteria such as accuracy, AUC, precision, recall, and F1-score, and these criteria are always expected to be higher.
However, these evaluation criteria are sensitive when there is a minority class.
At the same time, it is often difficult to choose the best model while the model exhibits the best result for some evaluation criteria, but not for all.
In order to address this problem, we consider MCDM, which considers the evaluation criteria of the mixer, some of which are expected to be higher, while others are expected to be lower.
Indeed, MCDM deals with various evaluation criteria and selects the best model.
In addition, previous studies have conducted experiments using a variety of experimental settings, such as selection of cross-validation techniques, up-sampling/down-sampling techniques, and hyperparameter optimization techniques, and did not provide any relative performance comparisons of different experimental settings to select the best model.
To solve the problem, we propose three training strategies under different experimental settings, and apply MCDM in each training strategy.
The MCDM results of each training strategy are integrated through ensemble methods to make the best decision for selecting the best model.

%Our work is as Cambridge works~\cite{1}, but the difference between our works is that we consider only spectral features while they consider spectral and deep features. 
%The reason for selecting only the spectral features is that the dataset is skewed and unbalanced.
%Deep learning models often overfit on such a small dataset, so we chose a different strategy. 
%Considering various spectral features, we use the ML model for classification.
%What's more, we integrated ensemble strategies in MCDM with several training strategies to make the best decision for choosing the best model.
%A model comparison that only considers one or two evaluation criteria cannot reflect the actual situation.
%Therefore, we consider MCDM that deals with various evaluation criteria, and to select the best model, we integrate ensemble methods in MCDM.
%Moreover, MCDM has proven its effectiveness in some aspects of the COVID-19 management system~\cite{54,55}. %To the best of our knowledge, this is the first attempt to explore ensemble-based MCDM in detecting COVID-19 from cough samples.
%Furthermore, the difference between our research and other research is that previous research did not attempt to explore cross-institutional validation, but we did.

%%%%% Proposed Architecture%%%%%%
\begin{figure}[!ht]
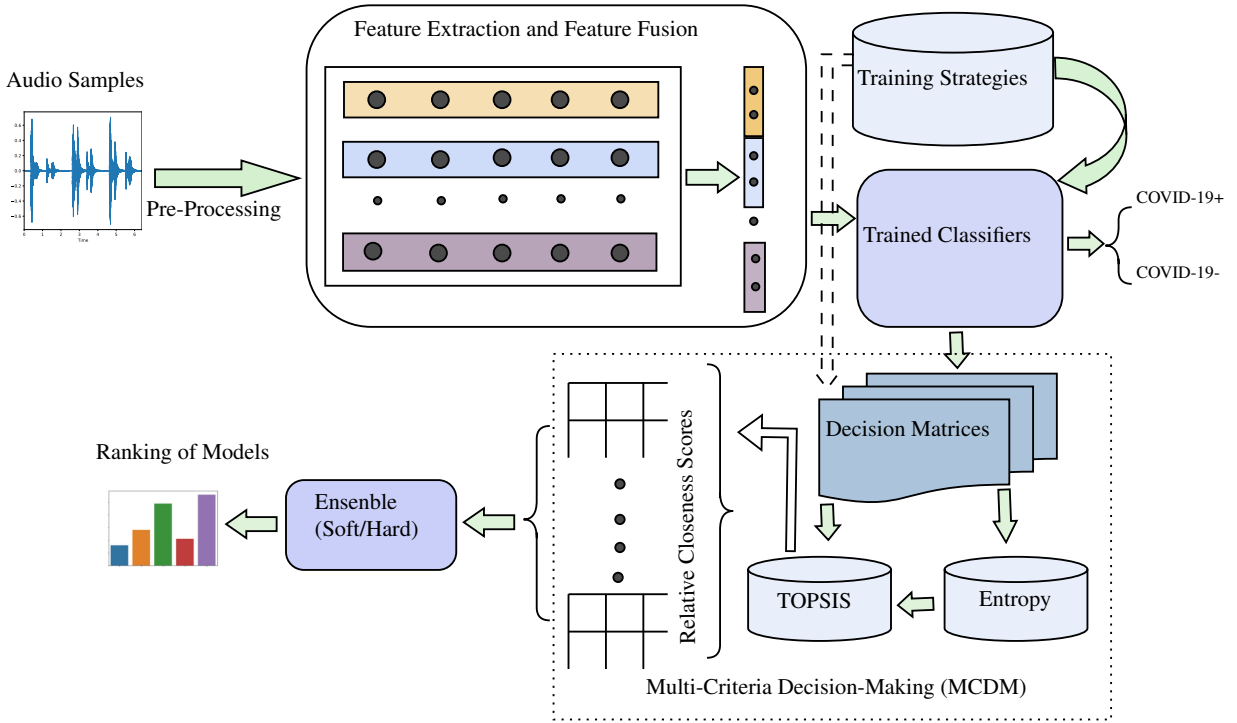

\centering
\resizebox{1\textwidth}{!}{

\tikzset{every picture/.style={line width=0.75pt}} %set default line width to 0.75pt        



}
\caption{An overview of the proposed method for detecting COVID-19 from cough samples.
}
\label{fig:COVID-19 cough detection system architecture}

\end{figure}

%%%%% Proposed Architecture%%%%%%

\section{Methodology}
\label{sec:Methodology}

Motivated by the current progress of ML-based audio applications, we have developed an end-to-end ML-based framework that can incorporate cough samples and directly predict binary classification labels, implying the possibility of COVID-19.
As the backbone of our proposed method, we use audio features, including Mel-Frequency Cepstral Coefficients, Mel-Scaled Spectrogram, Tonal Centroid, Chromagram and Spectral Contrast, and then perform feature fusion.
The output of the feature fusion passes to trained classifier layer that consists of $10$ classification methods, such as Extremely Randomized Trees (Extra-Trees), Support Vector Machine (SVM), Random Forest (RF), Adaptive Boosting (AdaBoost), Multilayer Perceptron (MLP), Extreme Gradient Boosting (XGBoost), Gradient Boosting (GBoost), Logistic Regression (LR), k-Nearest Neighbor (k-NN) and Histogram-based Gradient Boosting (HGBoost).
Each classifier is trained using different training strategies as detailed in Section~\ref{sec:Training strategies and hyper-parameters optimization}.
In addition, in order to select an optimized COVID-19 cough diagnosis model, we use the MCDM method that considers the decision matrix generated from different evaluation criteria  outlined in Section~\ref{sec:MCDM}. 
After that, we calculate the relative closeness score of each training strategy by integrating TOPSIS and entropy. 
Finally, we use two ensemble strategies (such as soft ensemble and hard ensemble) to rank the models. 
In the following sections, we outline the dataset description, the proposed method (including feature extraction and classification), the training strategies used, and the details of the optimization techniques used to select the best model.
An overview of our proposed method can be seen in Figure~\ref{fig:COVID-19 cough detection system architecture}.

\subsection{Dataset description and preprocessing}
\label{subsec:Dataset Description}

In this section, we will describe in detail the datasets used for analysis in this article.
We have used four datasets in the experimental evaluation: Cambridge~\cite{1}, Coswara~\cite{2}, Virufy~\cite{57}, and Virufy integrated with NoCoCoDa~\cite{17} datasets.
Table~\ref{tab:Dataset settings for validation} shows the distribution of cough samples used during experiment.
Each cough sample is resampled with a sampling rate of $22.5$ kHz, and a window type of Hann.

\begin{figure}[!ht]
  \centering
  \begin{subfigure}[b]{0.35\linewidth}
    \includegraphics[width=\linewidth,height=3cm]{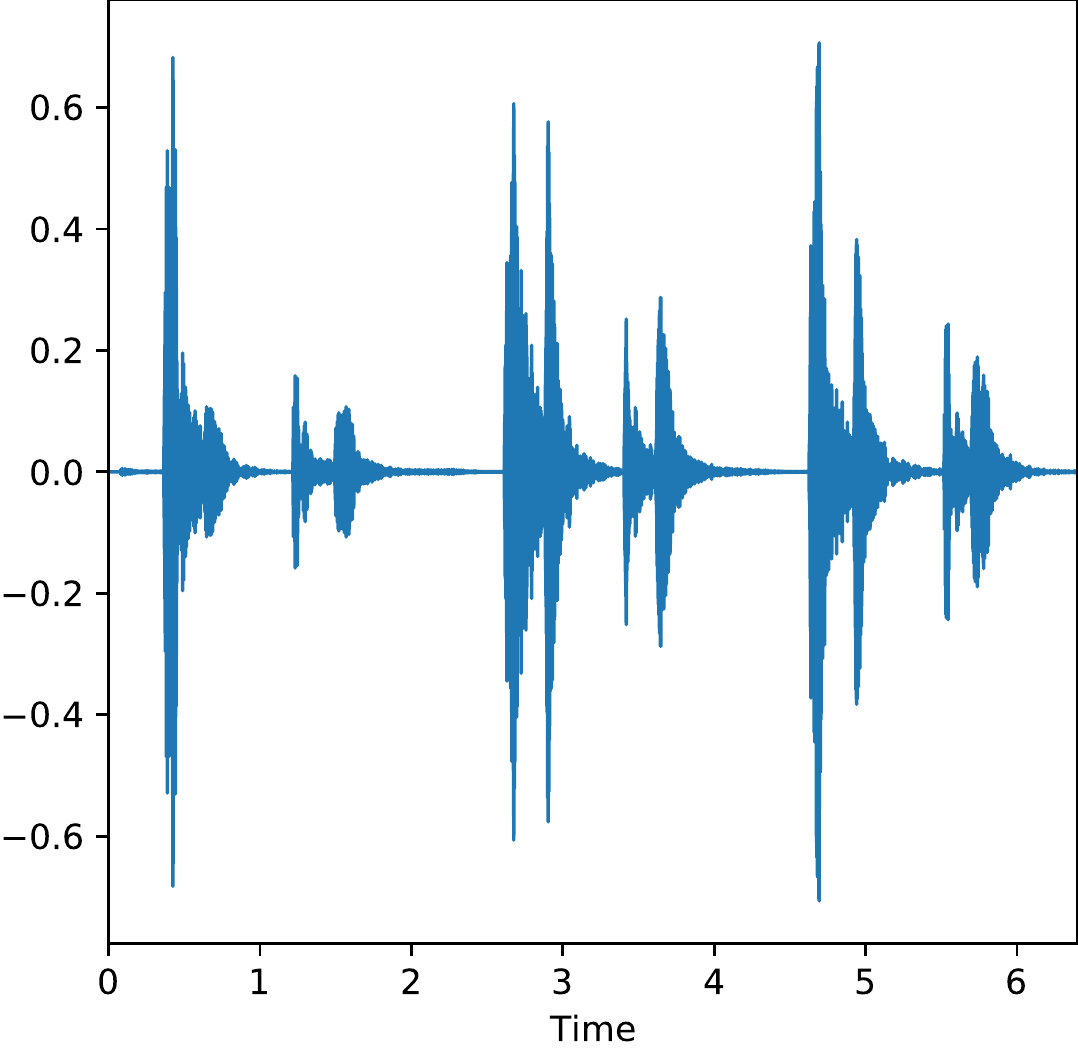}
    \captionsetup{justification=centering}
    \caption{COVID-19 Cough (Asymptomatic)}
  \end{subfigure}
  \begin{subfigure}[b]{0.35\linewidth}
    \includegraphics[width=\linewidth,height=3cm]{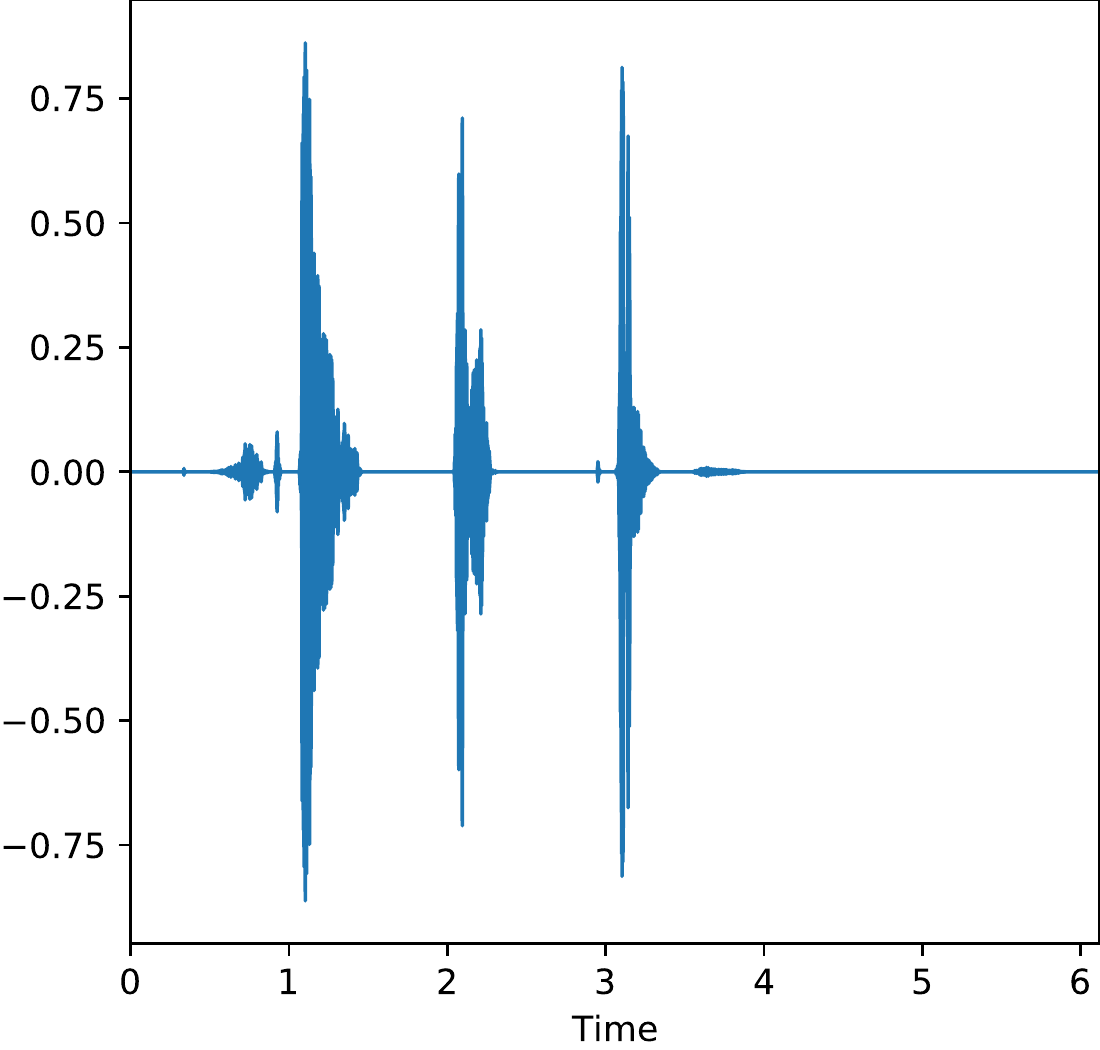}
    \captionsetup{justification=centering}
    \caption{Non-COVID-19 Cough (Asymptomatic)}
  \end{subfigure}
  \begin{subfigure}[b]{0.35\linewidth}
    \includegraphics[width=\linewidth,height=3cm]{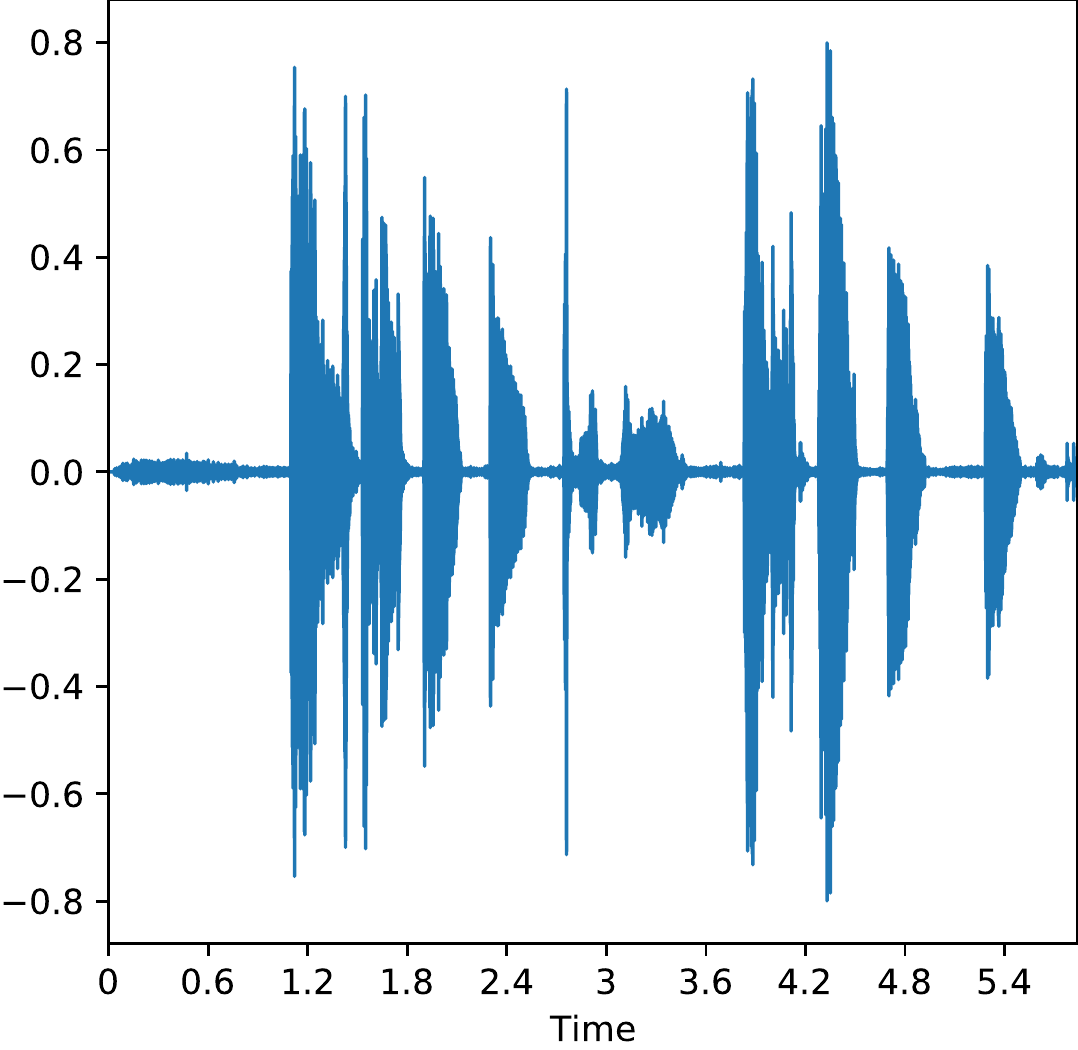}
    \captionsetup{justification=centering}
    \caption{COVID-19 Cough (Symptomatic)}
  \end{subfigure}
  \begin{subfigure}[b]{0.35\linewidth}
    \includegraphics[width=\linewidth,height=3cm]{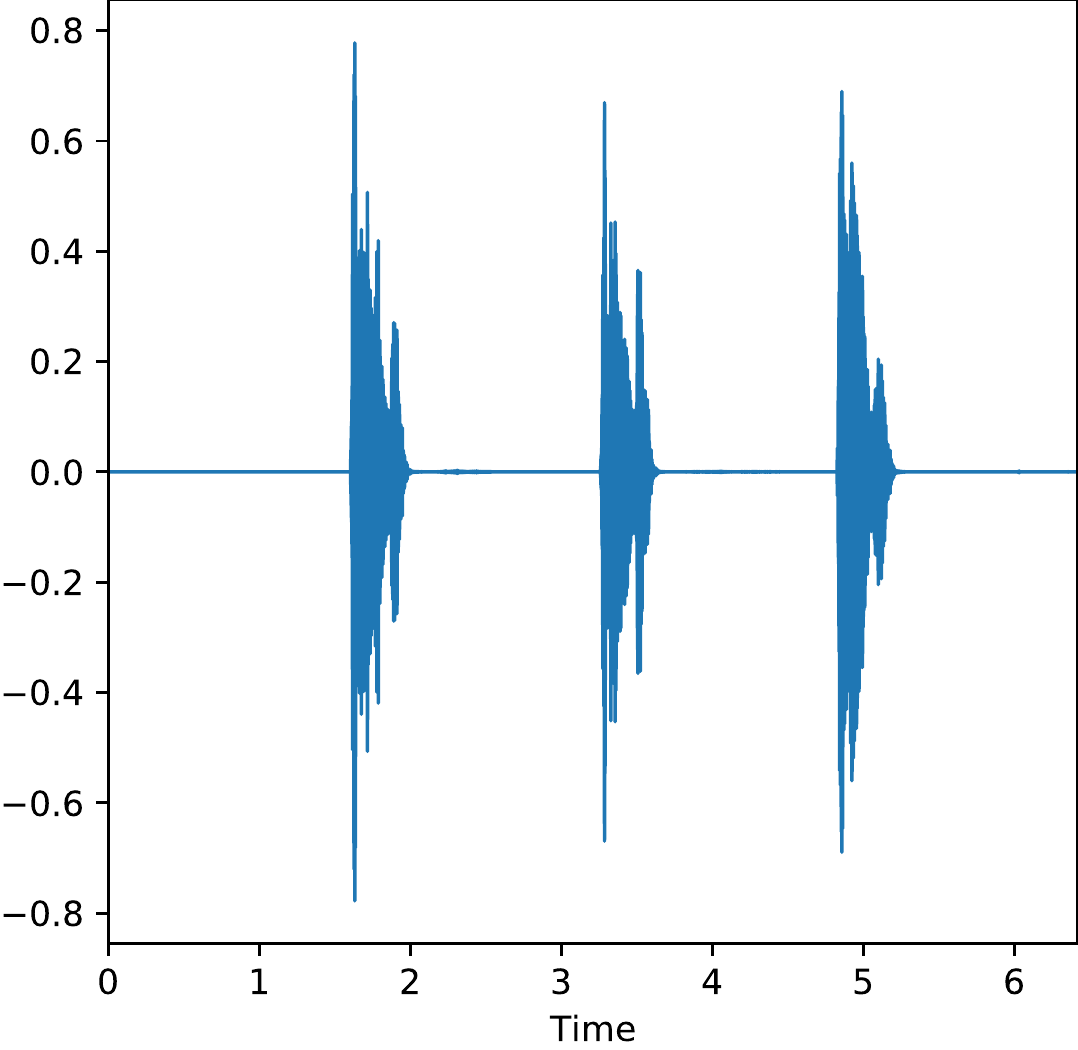}
    \captionsetup{justification=centering}
    \caption{Non-COVID-19 Cough (Symptomatic)}
  \end{subfigure}
  \captionsetup{justification=centering}
  \caption{COVID-19 and Non-COVID-19 cough samples of the Cambridge dataset}
  \label{fig:Cambridge Cough Sound Samples}
\end{figure}

\subsubsection{Cambridge dataset}
\label{subsubsec:Cambridge dataset}

The University of Cambridge has launched a web-based application and a mobile application for people to provide coughing, breathing, and voice when reading a prescribed sentence~\footnote{\url{https://www.covid-19-sounds.org/en/}}.
In the case of the Cambridge dataset, we consider two categories, namely asymptomatic and symptomatic, to distinguish COVID-19 positive from non-COVID-19.
Figure~\ref{fig:Cambridge Cough Sound Samples} shows asymptomatic and symptomatic COVID-19 and non-COVID-19 samples from the Cambridge dataset.
Since the authors of the University of Cambridge dataset released the dataset following a one-to-one legal agreement, we considered the restrictions they adopted to use it not for commercial purposes but for research purposes.

\begin{itemize}

 \item \textbf{Asymptomatic:} Distinguish people who tested positive for COVID-19 from those who tested negative, had a clean medical history, had never smoked, and were asymptomatic. In the dataset, there are $141$ cough samples from people who have tested positive for COVID-19 and $298$ cough samples from people who do not have COVID-19 (those who have a clean medical history, have never smoked, and have no symptoms). 

 \item \textbf{Symptomatic:} Distinguish between those who tested positive for COVID-19 and declared cough as a symptom from those who tested negative and had cough as a symptom. Moreover, these people had a clean medical history and had never smoked. This task distinguishes $54$ symptomatic COVID-19 samples from $32$ symptomatic non-COVID-19 samples.

\end{itemize}

\subsubsection{Coswara dataset}
\label{subsubsec:Coswara dataset}

In addition to the Cambridge dataset, we also consider the Coswara dataset that was developed by the Indian Institute of Science (IISc) Bangalore~\footnote{https://coswara.iisc.ac.in/}, and the dataset is now publicly available~\footnote{https://github.com/iiscleap/Coswara-Data}.
We collected samples from the Coswara dataset between April 2020 and May 2021. 
Since the record category of the Coswara dataset is different from that of the Cambridge dataset, in order to make it consistent with the Cambridge dataset, we only consider the heavy cough variants of the COVID-19 and healthy (non-COVID-19) categories.
From the Coswara dataset, we have considered a total of $185$ COVID-19 and $1,134$ non-COVID-19 cough samples for training and testing.

\subsubsection{Virufy dataset}
\label{subsubsec:Virufy dataset}

The Virufy COVID-19 open cough dataset is the first free COVID-19 cough sound, which is collected in the hospital under the supervision of a doctor in accordance with standard operating procedures (SOP) and the patient’s informed consent.
This dataset is preprocessed and labeled with COVID-19 status that is obtained through PCR testing and patient demographic data.
A total of $121$ segmented cough samples (including $48$ COVID-19 positive and $73$ COVID-19 negative) from $16$ patients were considered for experimental evaluation.

\subsubsection{NoCoCoDa dataset}
\label{subsubsec:NoCoCoDa dataset}

The NoCoCoDa dataset includes coughing events during or after the critical phase of COVID-19 patients recorded through public media interviews.
A total of $73$ individual cough events were obtained, and the cough phases were marked after the interview was manually segmented.
Since the NoCoCoDa dataset only has COVID-19 samples, in the experiment, we have integrated it with the Virufy dataset consisting of COVID-19 positive and healthy samples.

\begin{table}[!ht]
\normalsize
%\small
\centering
\caption{Datasets description}
\captionsetup{justification=centering}
%\resizebox{1\textwidth}{!}{
\begin{tabular}{llrrr} \hline
Dataset                    & Category    & COVID-19 & Non-COVID-19 & Total \\ \hline \hline
\multirow{2}{*}{Cambridge} & Asymtomatic & 141      & 298          & 439   \\ \cline{2-5}
                           & Symtomatic  & 54       & 32           & 86    \\ \hline
Coswara                    & -           & 185      & 1,134        & 1,319 \\ \hline
Virufy                     & -           & 48       & 73           & 121   \\ \hline
NoCoCoDa                   & -           & 73       & -            & 73    \\ \hline
Virufy+NoCoCoDa            & -           & 121      & 73           & 194  \\ \hline 
\end{tabular}
%}
\label{tab:Dataset settings for validation}
\end{table}

\subsection{Feature extraction methods}
\label{subsec:Feature Extraction Methods}

The sound waveform considered in the feature extraction process is sampled at a sampling rate of $22$kHz to ensure uniformity, as it is a standard frequency for audio applications.
From the sampled audio, five spectral features (i.e., Mel-Frequency Cepstral Coefficients, Mel-Scaled Spectrogram, Tonal Centroid, Chromagram, and Spectral Contrast) are extracted using the librosa~\cite{39} library from Python.

\begin{itemize}
  
  \item \textbf{Mel-Frequency Cepstral Coefficients (MFCCs):} MFCCs have already shown their usefulness through the analysis of dry and wet cough detection~\cite{40}, as well as highlighted as successful features for audio analysis. In the feature extraction of MFCC, after the windowing operation, fast fourier transform (FFT) applies to find the power spectrum of each frame. Afterward, the Mel scale is used to perform filter bank processing on the power spectrum. Mel-scaled filters are calculated from physical frequency ($f$) by the following equation~\ref{eqn:MFCCl}. After converting the power spectrum to the logarithmic domain, discrete cosine transform (DCT) is applied to the audio signal to measure the MFCC coefficients. 
  
  \begin{equation}
  \label{eqn:MFCCl}
   f_{mel}=2595\log_{10}\left ( 1+\frac{f}{700} \right )
  \end{equation}
  
  \item \textbf{Mel-Scaled Spectrogram:} In ML applications concerning audio analysis, we often require representing the power spectrogram in the Mel scale domain. The feature extraction process of the Mel-scaled Spectrogram includes several steps to generate the spectrogram. Before calculating the FFT, we set the window size to $2048$ and the hop length to $512$. After that, set the number of Mels to $128$, which is the evenly spaced frequency. Finally, the magnitude of the signal is decomposed into components corresponding to the frequencies in the Mel scale.
  
  \item \textbf{Tonal Centroid:} The tonal centroid feature is a way of projecting a $12$-bin tuned chromagram onto a $6$-dimensional vector, as described in equation~\ref{eqn:tonal centroid}~\cite{42}.  
  
  \begin{equation}
  \label{eqn:tonal centroid}
  \zeta _{n}\left ( d \right )=\frac{1}{\left \| c_{n} \right \|_{1}}\sum_{l=0}^{11}\Phi \left ( d,l \right )c_{n}\left ( l \right ),\quad  0\leq d< 5;\quad 0\leq l\leq 11
  \end{equation}
  
  where $\zeta _{n}$ is the tone centroid vector, and for the time frame $n$ is given by the product of the transformation matrix, $\Phi$, and the chroma vector $c$. 
  
  \item \textbf{Chromagram:} We calculate the chromatogram from the short-time Fourier transform (STFT) power spectrum. We initialize the window size to $2048$ and the hop length to $512$. The number of chroma bins generated is $12$. Finally, it extracts the normalized energy of each chroma bin on each frame, which is the required feature vector.
  
  \item \textbf{Spectral Contrast:} First, perform FFT on the audio samples to obtain the frequency spectrum. Using several Octave-scale filters, the frequency domain is partitioned into sub-bands. In the feature extraction process, the number of frequency bands is set to be $6$. The strength of spectral valleys, peaks, and their differences are evaluated in each sub-hand, as stated in equations (\ref{eqn:spectral contrast peak}, \ref{eqn:spectral contrast valley}, \ref{eqn:spectral contrast peak valley diff})~\cite{43}. After being converted to the logarithmic domain, the original spectral contrast features will be mapped to the orthogonal space.
  
  \begin{equation}
  \label{eqn:spectral contrast peak}
  Peak_{k}=log\left \{ \tfrac{1}{\alpha N}\sum_{i=1}^{\alpha N }x_{k,i} \right \}
  \end{equation}
  
  \begin{equation}
  \label{eqn:spectral contrast valley}
  Valley_{k}=log\left \{ \tfrac{1}{\alpha N}\sum_{i=1}^{\alpha N }x_{k,N-i+1} \right \}
  \end{equation}
  
  \begin{equation}
  \label{eqn:spectral contrast peak valley diff}
  SC_{k}=Peak_{k}-Valley_{k}
  \end{equation}
  
where $N$ is the total number in the $k$-th sub-band, $k\in[1,6]$, and $\alpha$ is a constant ranging from $0.02$ to $0.2$.
  
\end{itemize}

\begin{figure}[!ht]
  \centering
  
  \begin{subfigure}[b]{0.32\linewidth}
    \includegraphics[width=\textwidth]{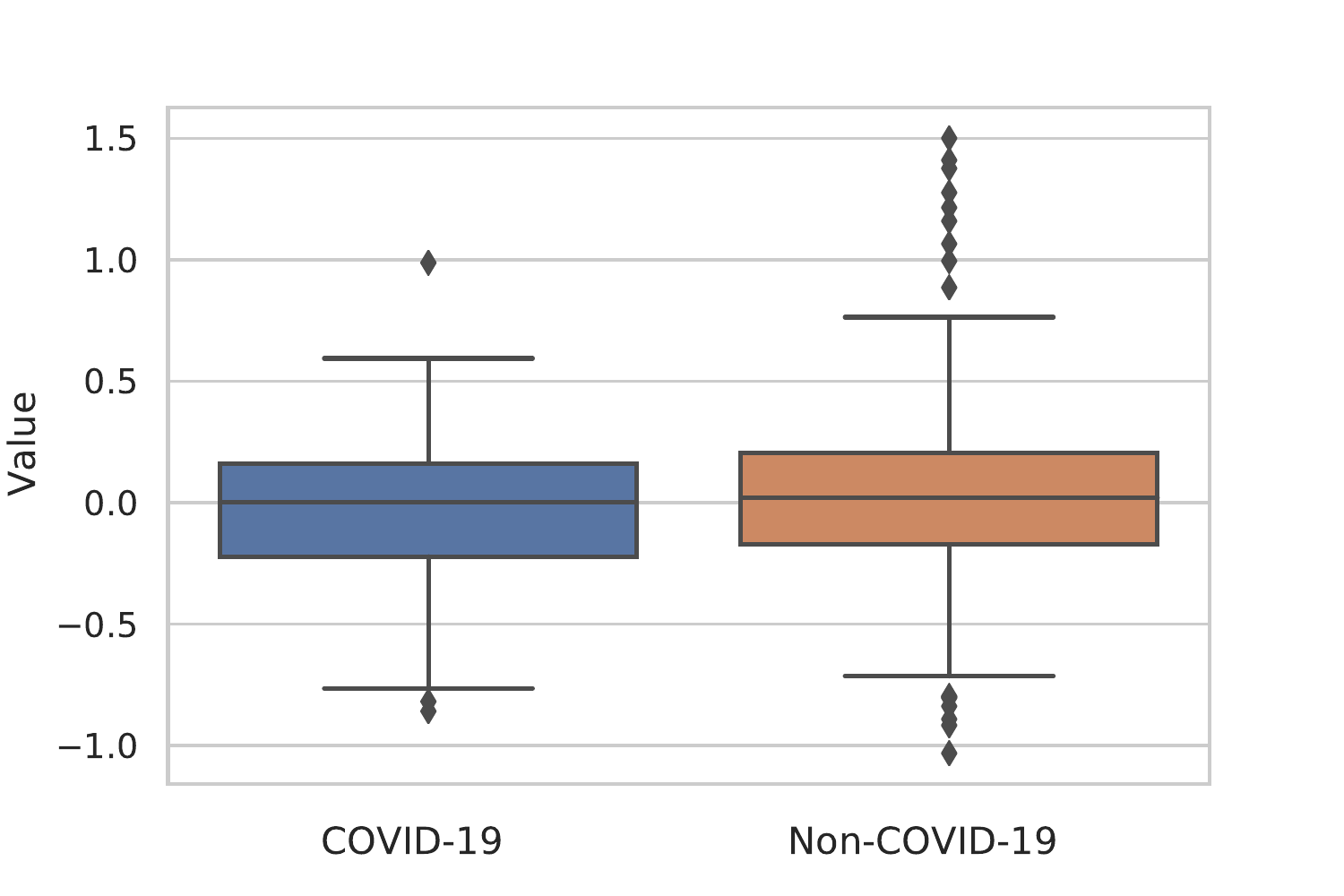}
    \captionsetup{justification=centering}
    \caption{MFCC}
  \end{subfigure}
  \begin{subfigure}[b]{0.32\linewidth}
    \includegraphics[width=\textwidth]{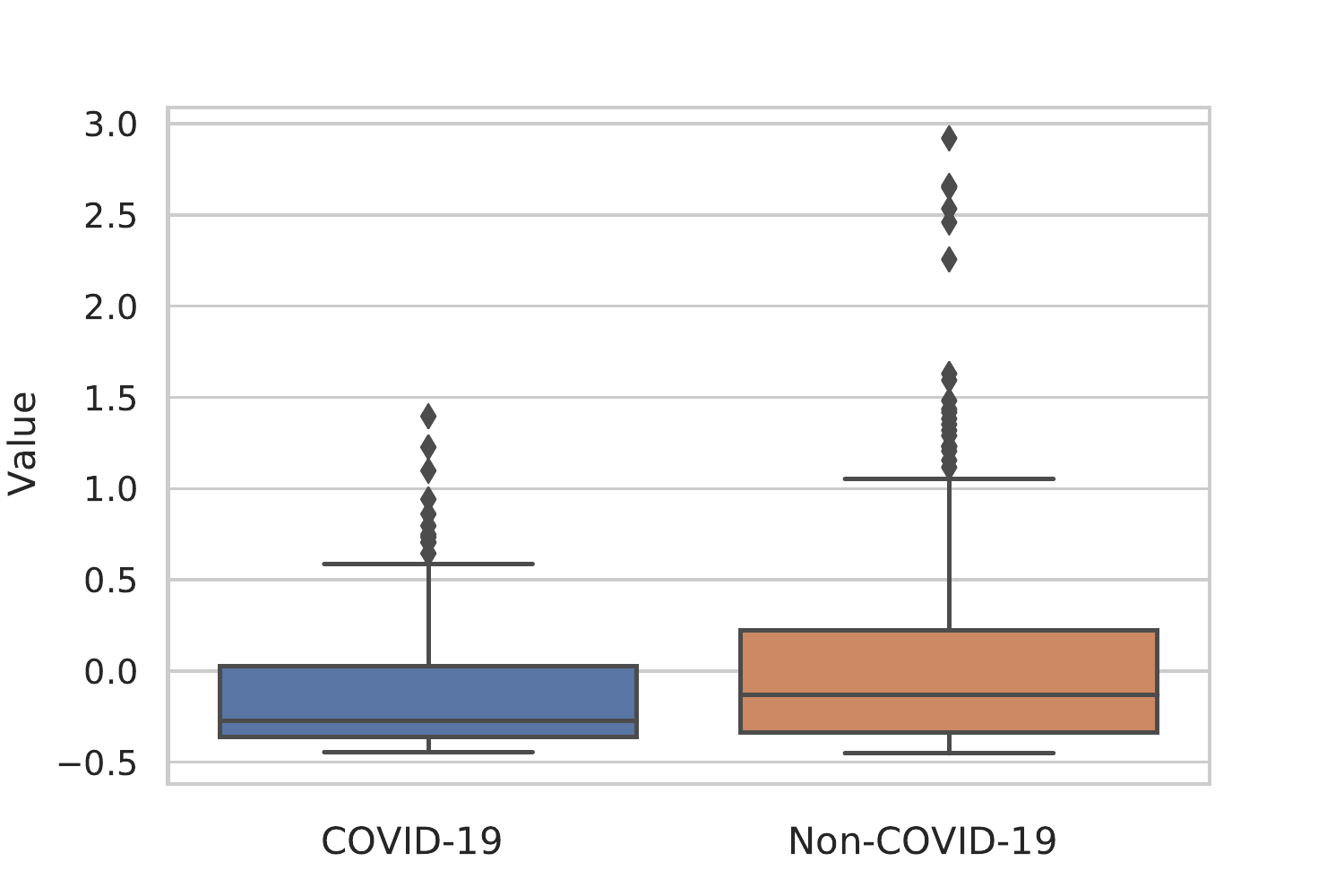}
    \captionsetup{justification=centering}
    \caption{Mel-scaled Spectogram}
  \end{subfigure}
  \begin{subfigure}[b]{0.32\linewidth}
    \includegraphics[width=\textwidth]{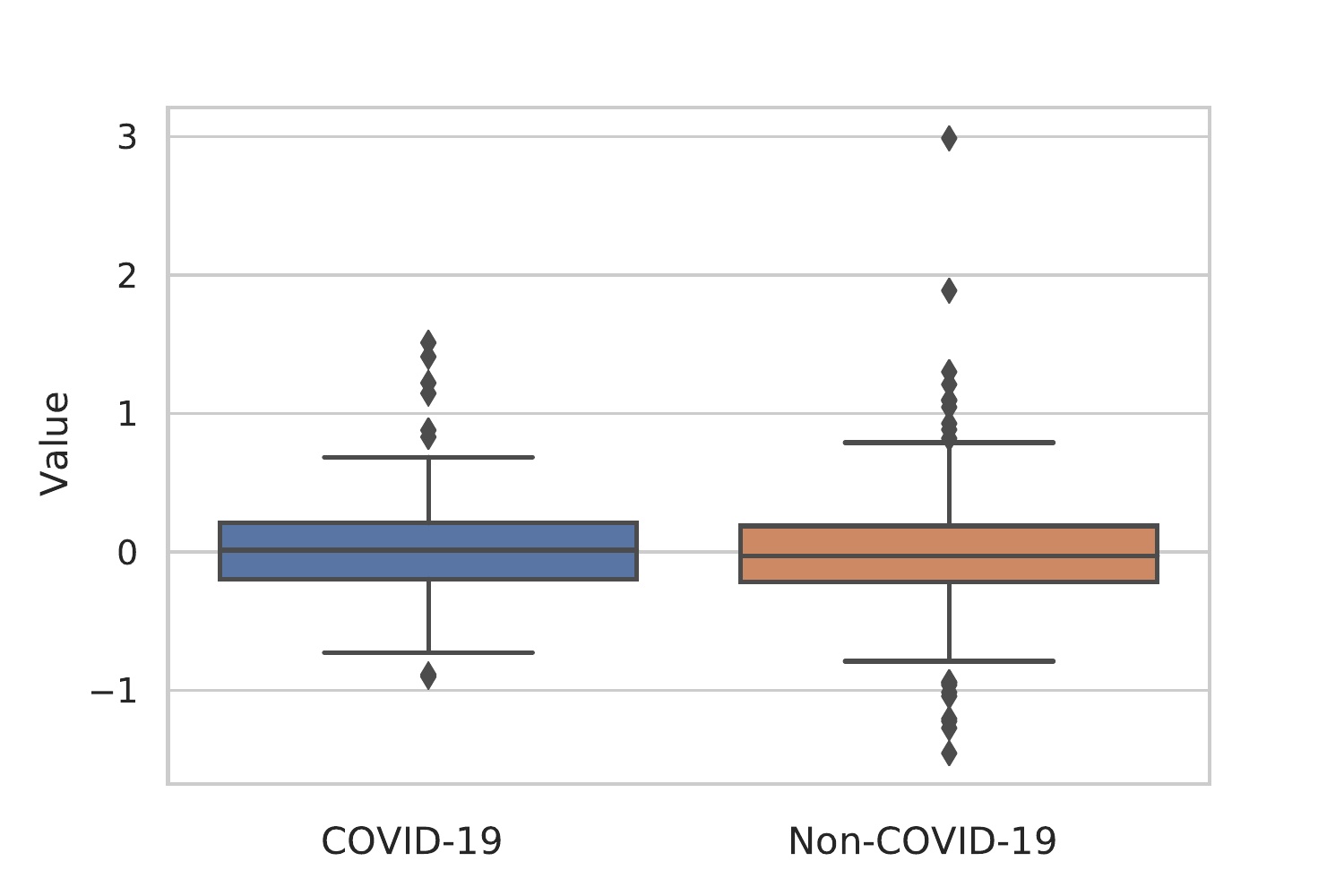}
    \captionsetup{justification=centering}
    \caption{Tonal Centroid}
  \end{subfigure}
  \begin{subfigure}[b]{0.32\linewidth}
    \includegraphics[width=\textwidth]{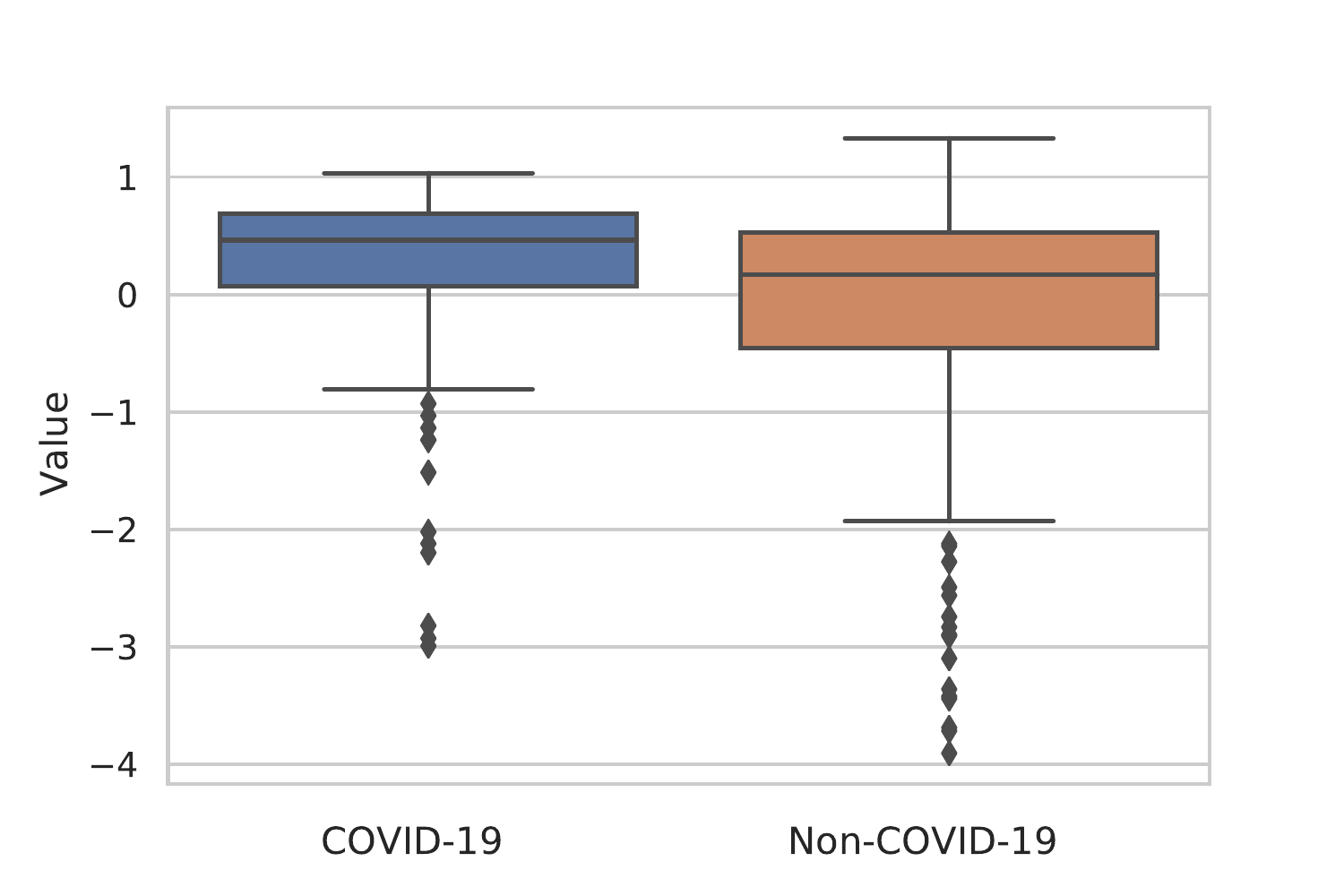}
    \captionsetup{justification=centering}
    \caption{Chromagram}
  \end{subfigure}
  \begin{subfigure}[b]{0.32\linewidth}
    \includegraphics[width=\textwidth]{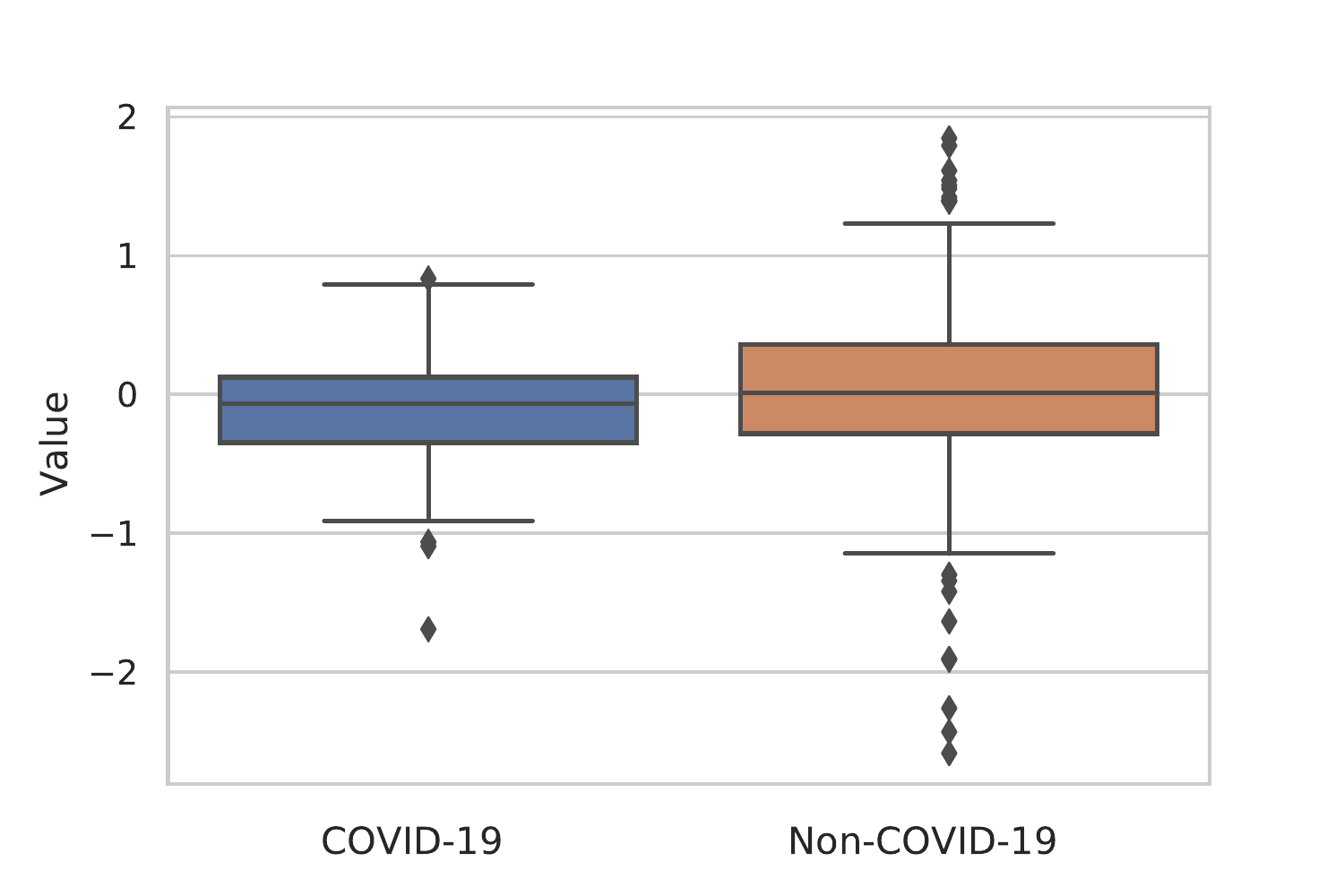}
    \captionsetup{justification=centering}
    \caption{Spectral Contrast}
  \end{subfigure}
   
  %\captionsetup{justification=centering}
  \caption{Box plots of asymptomatic cough samples for both COVID-19 and Non-COVID-19 in the Cambridge dataset.}
\label{fig:Box plot sound featutre}
\end{figure}

\begin{figure}[!ht]
    \centering
    \includegraphics[width=.6\textwidth]{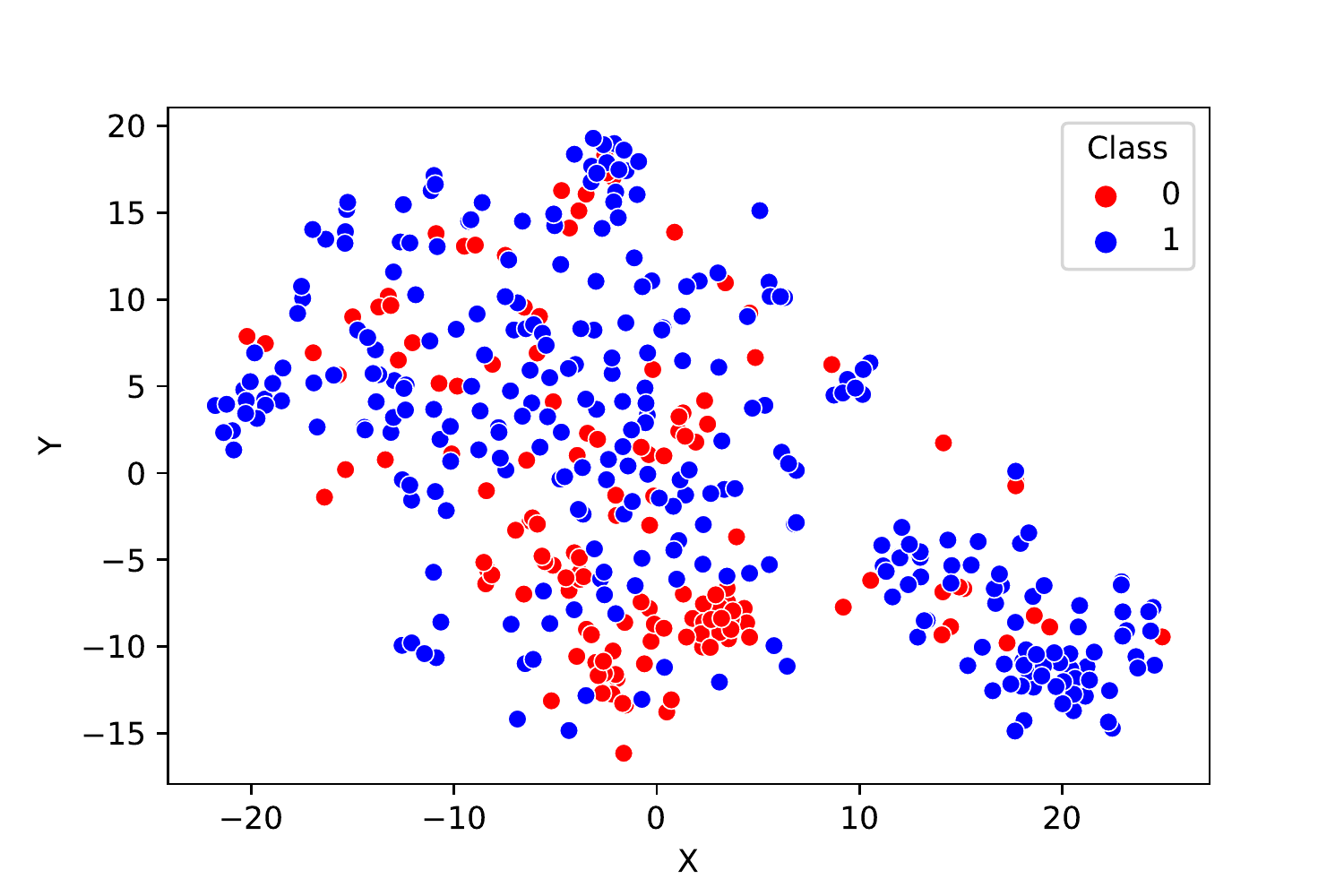}
    \caption{Use tSNE to visualize asymptomatic COVID-19 and Non-COVID-19 cough samples in the Cambridge dataset. Note that 0 represents COVID-19 and 1 represents Non-COVID-19 cough.}
\label{fig:tSNE_feature_Cambridge}
\end{figure}

The features generated using the above methods can be obtained by adjusting several hyper-parameters.
There are a total of $193$ audio features, including $40$ components of MFCC, $128$ components of Mel-scaled Spectrogram, $12$ components of Chromagram, $7$ components of Spectral Contrast, and $6$ components of Tonal Centroid. 
Figure~\ref{fig:Box plot sound featutre} shows box plots of the mean features of asymptomatic cough samples for COVID-19 and non-COVID-19 in the Cambridge dataset.
In the chromatogram, because the median is close to the upper quartile, the features are negatively skewed. On the other hand, for Mel-scaled Spectogram, features are positively skewed. For the other three feature types (i.e., MFCC, Spectral Contrast, Tonal Centroid), the features are distributed symmetrically.
Samples from COVID-19 are more concentrated on the mean of the distribution, while non-COVID-19 samples maintain a wider range, reflecting more scattered data.
Next, Figure~\ref{fig:tSNE_feature_Cambridge} depicts a two-dimensional visualization of asymptomatic cough sample features for the two classes (i.e., COVID-19 and Non-COVID-19) in the Cambridge dataset through t-distributed Stochastic Neighbor Embedding (t-SNE)~\cite{44}.
As the features of the cough audio samples are multi-dimensional, t-SNE (a non-linear dimensionality reduction technique) is used, because it is very suitable for visualizing multi-dimensional data in a low-dimensional space.

\subsection{Trained classifiers}
\label{subsec:Classification Methods}

%We can solve the problem of COVID-19 cough detection from the ML aspect.
We consider ten ML algorithms in our proposed method for classification, i.e., Extremely Randomized Trees (Extra-Trees), Support Vector Machine (SVM), Random Forest (RF), Adaptive Boosting (AdaBoost),  Multilayer Perceptron (MLP), Extreme Gradient Boosting (XGBoost), Gradient Boosting (GBoost), Logistic Regression (LR), k-Nearest Neighbor (k-NN) and Histogram-based Gradient Boosting (HGBoost).
In the following, we will briefly describe each of the different classifiers evaluated in our experimental evaluation.

\begin{itemize}

\item \textbf{Extremely Randomized Trees (Extra-Trees)} is a classifier that can fit multiple random decision trees to each sub-sample of the dataset, so it can control over-fitting and usages the average to improve detection accuracy. 
Extra-Trees classifier has proven to be useful in diagnosing patients with chronic obstructive pulmonary disease~\cite{45}.

\item \textbf{Support Vector Machine (SVM)} is a popular supervised technique that can effectively perform classification tasks. Several SVM kernels (such as Gaussian function, polynomial function, or quadratic function) can be used during the classification task. Some previous studies (\cite{18,1,4,22,41,47,49}) have successfully applied SVM to detect COVID-19 in audio samples. 

\item \textbf{Random Forest (RF)} is a collection of decision trees, which is widely used in classification tasks. By growing a combination of trees and voting for each category of trees, we can observe significant classification accuracy. Random forest has achieved success in classifying cough, breath, and sound events~\cite{2}. 

\item \textbf{Adaptive Boosting (AdaBoost)} is a classifier that first fits the classifier to the original dataset, and then fits other copies of the classifier fit the same dataset. However, the weights of misclassified instances are adapted to force successive classifiers to pay more attention to hard events.

\item \textbf{Multilayer Perceptron (MLP)} has adapted to the concept of human biological neural networks and can learn non-linear relationships. The training of the network depends on iteration, bias, weight adjustment, learning rate,  and optimization. It is effective in detecting COVID-19 coughs~\cite{37,41,47} and other types of coughs.

\item \textbf{Extreme Gradient Boosting (XGBoost)} classifier is a decision-tree-based ensemble ML technique that utilizes a gradient boosting structure. This is an advanced and powerful technique that can deal with data irregularities and further reduce overfitting~\cite{49}. Some previous studies have reported the performance of the XGBoost classifier in detecting COVID-19 in cough samples~\cite{4,49}. 

\item \textbf{Gradient Boosting (GBoost)} generates an additive model according to the forwarding stage-wise, and summarizes it by optimizing the differentiable loss function~\cite{33}. At each stage, regression trees (equal to the total number of classes) are fitted to the negative gradient of the binomial or multinomial deviation loss function.

\item \textbf{Logistic Regression (LR)} is a parametric classification model with fixed parametric numbers that predict categorical or discrete output for given input features. We can use multinomial logistic regression in scenarios with multiple categories rather than two categories~\cite{11}. Madhurananda et al.~\cite{41} has successfully used it for COVID-19 cough detection. 

\item \textbf{k-Nearest Neighbor (k-NN)} is a well-known classifier that appears in large-scale ML applications. As we have seen from previous studies, researchers not only used k-NN in non-COVID-19 applications such as night coughing and sniffing~\cite{46} but also used k-NN in the detection of COVID-19 in cough samples~\cite{41,47,49,48}. 

\item \textbf{Histogram-based Gradient Boosting (HGBoost)} is a highly desirable ML technology, where the application needs to get better quality performance in less inference time. The main advantage of histogram-based gradient boosting technology is speed. Chung et al.~\cite{50} has successfully explored this method to predict the severity of COVID-19. 

\end{itemize}

\subsection{Training strategies and hyper-parameters optimization}
\label{sec:Training strategies and hyper-parameters optimization}

We introduce three training strategies, namely training strategies $1$, $2$ and $3$, to evaluate the effectiveness of different factors of the proposed method.
It is obvious from the dataset that the positive category of COVID-19 is under-represented, which may adversely affect the performance of the ML classifier.
Therefore, we have used the Synthetic Minority Oversampling Technique (SMOTE)~\cite{36} during training to balance the dataset to enhance the ML classifier’s performance.
The difference between training strategy $1$ and strategy $2$ is that strategy $1$ does not apply
SMOTE in the training process, while strategy $2$ does.
However, they both use the same hyper-parameters.
On the other hand, the difference between strategies $1$ and $2$ and strategy $3$ is that strategy $3$ integrates nested cross-validation with hyper-parameters optimization.
The nested cross-validation includes an inner loop of $5$-fold stratified cross-validation for hyper-parameters optimization, and the outer loop, being in the training process with SMOTE, maintains $10$-fold stratified cross-validation.
The hyper-parameters used during empirical evaluation for optimization are listed in Table~\ref{Hyper-parameter configuration}.
For classifications where we encounter class imbalance problems, the default threshold (i.e, 0.50) leads to poor performance. 
Therefore, we apply the threshold moving technique to adjust the probability threshold that outlines the probability to the class label.
In the experiment, we define a set of thresholds, and then measure the predicted probability of each threshold to select the best threshold.
Table~\ref{Configurations of training Strategies} shows different configurations of training strategies.

\begin{table}[!ht]
\small
\centering
\caption{Hyper-parameters search space of classifiers for optimization}
\captionsetup{justification=centering}
%\resizebox{1\textwidth}{!}{
\begin{tabular}{lll}
\hline
Classifiers                  & Hyper-parameters   & Range       \\ \hline \hline
\multirow{3}{*}{Extra-Trees} & Estimators         & 600, 700, 800                            \\
                             & Criterion          & Gini, Entropy                          \\
                             & Max. features      & Auto, Sqrt, Log2                       \\ \hline
\multirow{3}{*}{SVM}         & C                  & 0.10 to 1.0, step=0.10                 \\
                             & Kernel             & Linear, Poly, rbf, Sigmoid             \\
                             & Gamma              & Auto, Scale                            \\ \hline
\multirow{2}{*}{RF}          & Estimators         & 600, 700, 800                            \\
                             & Max. features      & Auto, Sqrt, Log2                       \\ \hline
\multirow{2}{*}{AdaBoost}    & Estimators         & 600, 700, 800                            \\
                             & Algorithm          & SAMME, SAMME.R                         \\ \hline
\multirow{4}{*}{MLP}         & Hidden layer sizes & (64), (64,64), (128), (128,128)           \\
                             & Activation         & identity, logistic, tanh, relu         \\
                             & Solver             & lbfgs, sgd, adaml                      \\
                             & Learning rate      & constant, invscaling, adaptive         \\ \hline
\multirow{2}{*}{XGBoost}     & Estimators         & 600,700,800                            \\
                             & Max. depth         & 4,5,6 \\  \hline  
\multirow{4}{*}{GBoost}      & Estimators         & 600, 700, 800                            \\
                             & Criterion          & friedman\_mse, mse                     \\
                             & Max. features      & auto, sqrt, log2                       \\
                             & Loss               & deviance, exponential                  \\ \hline
\multirow{2}{*}{LR}          & Penalty            & l1, l2, elasticnet                     \\
                             & Solver             & newton-cg, lbfgs, liblinear, sag, saga \\ \hline
\multirow{2}{*}{k-NN}        & Number of neighbours & 5 to 8, step=1                         \\
                             & Algorithm          & auto, ball tree, kd tree, brute        \\ \hline
\multirow{2}{*}{HGBoost}     & Max. iteration     & 100 to 600, step=100                   \\
                             & Loss               & binary crossentropy                    \\ \hline
                               
\end{tabular}

%}
\label{Hyper-parameter configuration}
\end{table}

\begin{table}[!ht]
%\normalsize
\large
\centering
\caption{Configurations of different training strategies}
\captionsetup{justification=centering}
\resizebox{1\textwidth}{!}{
\begin{tabular}{cccccc}
\hline
\makecell[l]{Training\\Strategy \#} & \makecell{Cross-Validation \\ Method} & \makecell{Cross-Validation  \\Folds} & \makecell{Up-sampling \\ Method} & Threshold Moving & \makecell{Hyper-parameters \\Selection Method}                                        \\ \hline \hline
Strategy 1            & Stratified                 & 10                       & N/A                   & \checkmark              & Fixed                                                    \\ \hline
Strategy 2            & Stratified                 & 10                       & SMOTE                 & \checkmark              & Fixed                                                    \\ \hline
Strategy 3            & Stratified                 & 10                       & SMOTE                 & \checkmark              & \makecell{Optimized using Nested \\ Cross-Validation with Grid Search} \\ \hline 

\hline
\end{tabular}
}
\label{Configurations of training Strategies}
\end{table}

\subsection{Ensemble-based MCDM}
\label{sec:MCDM}

Towards the selection of an optimized COVID-19 cough diagnostic model, we have employed the MCDM method that considers different evaluation criteria.
Selecting the best model using one or few evaluation criteria (such as accuracy, precision, etc.) does not make proper sense when we consider bias data, i.e., class imbalance, where most data belongs to one class.
To address this problem, we consider MCDM, which considers several evaluation criteria with higher and lower influence in the mixer.
For example, some evaluation criteria are expected to have high values, such as accuracy, precision, etc., while we expect other evaluation criteria to have low values, such as false positive rate, false negative rate, etc.
One widely accepted approach for MCDM is the integration of Entropy and TOPSIS methods where Entropy calculates the weight of each evaluation criterion and TOPSIS handles this weight with a decision matrix to produce an outcome that reflects the best performing model.
TOPSIS has the following advantages: (1) Suitable for processing many alternatives and attributes; (2) The process is simple and easy to use; (3) Regardless of the number of attributes, it maintains the same processing steps~\cite{52}.
The core aspect of the TOPSIS method is the decision matrix, which is formed by using the evaluation criteria value of each alternative, as defined in Equation~\ref{eqn:DM}.

\begin{equation}
  \label{eqn:DM}
   D=\bordermatrix{\text{}&C_1&C_2&\ldots &C_n\cr
                A_1&X_{11} &  X_{12}  & \ldots & X_{1n}\cr
                A_2& X_{21}  &  X_{22} & \ldots & X_{2n}\cr
                \vdots& \vdots & \vdots & \ddots & \vdots\cr
                A_m& X_{m1}  &   X_{m2}       &\ldots & X_{mn}}
  \end{equation}

Where, $A_1,A_2,...,A_m$ represent the alternatives to ranking based on the evaluation criteria and $C_1,C_2,...,C_n$. $X_{ij}$ represents the score of the alternative $A_i$ related to the criterion $C_j$. 

Entropy-based weight measures the information of the decision matrix, which is the prerequisite of the TOPSIS method development, and is used to determine the criterion’s weight. 
We not only use entropy to quantitatively measure data, but also calculate proportional weight information. 
We have summarized the complete working steps of determining the weight of each evaluation criterion in Algorithm~\ref{entropy}.
Supposing there are $m$ alternatives and $n$ pieces of criteria in the $D$, $X_{ij}$ is the $j$-th criterion value in the $i$-th alternative. 
The algorithm includes several steps: the standardization of the index, the element-wise projection, measurement of entropy of the $j$-th index, and calculation of weight of each criterion.

\begin{algorithm}[!htb]
	\caption{Steps to measure entropy-based weight}
	\label{entropy}
\begin{algorithmic}[1]
	\STATE \textit{Input:}  Decision matrix, $\mathbf{D}=\left [ X_{ij} \right ]_{m\times n}$
	\STATE \textit{Output:} Evaluation criteria weights, $W_{j}$
	\STATE ${X}_{ij}'\gets \frac{X_{ij}-\underset{j}{\min}\,X_{ij}}{\underset{j}{\max}\,X_{ij}-\underset{j}{\min}\,X_{ij}} \: \: \forall i,j$ \Comment*[r]{Standardization of indexes}
	\STATE ${X}''_{ij} \gets \frac{{X}'_{ij}}{\sum_{i=1}^{m}{X}'_{ij}} \: \: \forall i,j$ \Comment*[r]{Projected result after standardization}
    \STATE$E_{j} \gets \frac{-1}{\ln m}\sum_{i=1}^{m}{X}''_{ij}\ln \left ( {X}''_{ij} \right ) \: \: \forall j$ \Comment*[r]{Entropy of the $j$th index}
    \STATE $W_{j} \gets \frac{1-E_{j}}{\sum_{j=1}^{n}1-E_{j}} \forall j$ \Comment*[r]{Entropy weight of the $j$th index}
	\STATE $Return$ $weights$
\end{algorithmic}
\end{algorithm}

\begin{algorithm}[!t]
	\caption{Steps of TOPSIS method}
	\label{TOPSIS}
\begin{algorithmic}[1]
	\STATE \textit{Input:} Decision matrix, $\mathbf{D}=\left [ X_{ij} \right ]_{m\times n}$
	\STATE \textit{Output:} Rank of each model, $R_{i}$
	\STATE $X^{N}_{ij} \gets \frac{X_{ij}}{\sqrt{\sum_{i=1}^{m}X^{2}_{ij}}} \: \: \forall j$ \Comment*[r]{Normalized decision matrix} 
	\STATE $V_{ij}\gets W_{j}X^{N}_{ij} \: \: \forall i,j $ (see Algorithm~\ref{entropy}) \Comment*[r]{Weighted normalized decision matrix}
	\STATE $V^{+}\gets\left ( v^{+}_{1},v^{+}_{2},...,v^{+}_{n} \right ),\: \: V^{-}\gets\left ( v^{-}_{1},v^{-}_{2},...,v^{-}_{n} \right ) $ \Comment*[r]{Determine ideal best and ideal worst solution}
	\STATE $S^{+}_{i}\gets\sqrt{\sum_{j=1}^{n}\left ( V_{ij}-v^{+}_{j} \right  ) ^{2}},\: \: S^{-}_{i}\gets\sqrt{\sum_{j=1}^{n}\left ( V_{ij}-v^{-}_{j} \right )^{2}}\: \: \forall i$ \Comment*[r]{Measure separation}
	\STATE $C_{i} \gets \frac{S^{-}_{i}}{S^{+}_{i}+S^{-}_{i}}, \: \: 0\leq C_{i}\leq 1 \: \: \forall i$ \Comment*[r]{Measure relative closeness}
	\STATE $R_{i}\gets Rank\left ( C_{i} \right ) \: \: \forall i$ \Comment*[r]{Ranking of relative closeness}
	\STATE $Return$ $closeness\;score\;and\;ranks$
\end{algorithmic}
\end{algorithm}

We outline the functional steps of the TOPSIS method in Algorithm~\ref{TOPSIS}.
After performing the initial steps of the TOPSIS, i.e., normalization of the decision matrix and determination of the weighted decision matrix, Step 3 in Algorithm~\ref{TOPSIS} defines the ideal best and the ideal worst solution.
The equations for determining the ideal best and the ideal worst are as follows:

\begin{equation}
  \label{eqn:ideal best}
   V^{+} =\left \{ \left ( \underset{j}{\max}\,V_{ij}\mid j\in J_{+} \right ),\left (\underset{j}{\min}\,V_{ij}\mid j\in J_{-} \right ); \: \: \forall i\right \} 
  \end{equation}
 
\begin{equation}
  \label{eqn:ideal worst}
   V^{-} =\left \{ \left ( \underset{j}{\min}\,V_{ij}\mid j\in J_{+} \right ),\left (\underset{j}{\max}\,V_{ij}\mid j\in J_{-} \right ); \: \: \forall i\right \}  
  \end{equation}

where $J_{+}$ and $J_{-}$ are the criteria having positive and negative impact respectively.
Step 4 calculates the distance between each feasible solution and the ideal positive solution and the ideal negative solution.
Next, step 5 measures the relative closeness to the ideal solution, and finally, step 6 ranks the evaluation alternatives according to the relative closeness value.

\begin{algorithm}[!t]
\caption{Steps of soft and hard ensemble method during validation}
\label{ensemble}
\begin{algorithmic}[1]

	\STATE \textit{Input:} Training and testing samples; ML models
	\STATE \textit{Output:} Best models using soft and hard ensemble
	
    \FOR{$t \gets 1$ to $N$}
    \FOR{$m \gets 1$ to $M$}
	    \STATE Measure optimized parameter,$Q_t$ for    training strategy $t$ of model $m$
	    \STATE Predict test samples using, $Q_t$
            Create decision matrix, $D$
        \STATE Measure MCDM relative closeness, $C_{mt}$ (see Algorithm~\ref{TOPSIS})
	 \ENDFOR
	 \ENDFOR
 
	\STATE $S_{i} \gets \frac{1}{T}\sum_{j=1}^{T}C_{ij} \: \:\forall i$ \Comment*[r]{Calculate soft ensemble score} 
	\STATE $R_{i} \gets rank\left ( S_{i}  \right )\: \:\forall i; \: \: S_{best}\gets \underset{i}{\max}\,R_{i}$ \Comment*[r]{Measure rank and best model (soft ensemble)}
	\STATE $H_{i} \gets \sum_{j=1}^{T}rank\left ( C_{ij}  \right )\: \:\forall i; \: \: H_{best}\gets \underset{i}{\max}\,H_{i} $ \Comment*[r]{Calculate hard vote and choose best model (hard ensemble)}
	\STATE $Return$ $model's\;rank$

\end{algorithmic}
\end{algorithm}

We also integrate ensemble methods into MCDM in combination with the multiple training strategies discussed in Section~\ref{sec:Training strategies and hyper-parameters optimization}.
The core concept of multiple training strategies is developed on the basis of considering the training strategies of different experimental settings.
Each experimental setup contains unique optimization parameters.
Therefore, ensemble in MCDM through multiple training strategies is more efficacious than MCDM based on one training, thus providing a better model choice when diagnosing COVID-19 cough.
We have selected two ensemble methods in the proposed method, such as \emph{soft ensemble} and \emph{hard ensemble}, to select the best model in MCDM to classify cough samples as COVID-19 or non-COVID-19, which has been carried out in Algorithm~\ref{ensemble} description.
In Algorithm~\ref{ensemble}, steps $1$ to $7$ have measured the relative closeness of MCDM of each model for each training strategy. 
With a soft ensemble, it uses the average value of relative affinity and considers all training strategies, and ranks the models according to the average value, as described in steps $9$ and $10$. 
Using a hard ensemble, the outcome of an MCDM is defined as the transformation of relative closeness score that maps to a vote. 
The final ensemble needs to aggregate the votes of all training strategies for all alternatives (i.e., classification models), and select the best alternative category with the highest number of votes, as shown in step $11$.

\section{Experiments and Results}
\label{sec:resultanalysis}

In this section, we presents our experimental results to detect COVID-19 from cough sound.
We first describe the evaluation criteria used in experimental evaluation (Section~\ref{subsec:Evaluation metrics}). 
After that, using the Cambridge dataset, we present the classification performance of our approach (Sections~\ref{subsec:Prediction performance of Task 1} and \ref{subsec:Prediction performance of Task 2}), and the ranking of the classification models using ensemble-based MCDM (Section~\ref{subsec:Model selection using ensemble-based MCDM}).
Then, we discuss the feature selection process using Recursive Feature Elimination with Cross-Validation (RFECV) method and apply this process to all datasets used in this experiment.
%Table~\ref{tab:Comparision result with feature reduction} shows the results of this step.
Finally, we present a comparison of our approach with the state-of-the-art approaches and show the results of other datasets.

\subsection{Evaluation criteria}
\label{subsec:Evaluation metrics}

We use eight standard evaluation metrics such as Accuracy (Acc.), Receiver Operating Characteristic - Area Under Curve (ROC-AUC), Precision, Recall, Specificity, F1-score, False Positive Rate (FPR), False Negative Rate (FNR) across all $10$-fold stratified cross-validation. 
%In addition, we measure the $p$-value, which is a quantitative method for reporting statistical hypothesis test results. 
%Contrasted with specific performance indicators, $95\%$ CI, which is a more efficacious indicator, is also considered.
%It can show the reliability of the problem domain and increase statistical significance. 
%We further present the mean and standard deviation of accuracy (Acc.) and ROC-AUC across all $10$-fold stratified cross-validation.

\subsection{Prediction performance of asymptomatic category}
\label{subsec:Prediction performance of Task 1}

We present the decision matrix related to the classification performance of the various classifiers in Table~\ref{tab:Task 1 result} for asymptomatic category of the Cambridge dataset.
The evaluation criterion linked to the upward arrow expects to have a higher value, while the downward arrow is the opposite.
For training strategy 1, the results indicate that the Extra-Trees classifier provides best performance, with AUC, accuracy, precision, and recall of $0.85$, $0.86$, $0.93$, and $0.62$, respectively.
In addition, HGBoost and RF classifiers also show excellent performance, with AUC of $0.83$ and $0.81$, respectively. 
However, XGBoost classifier manifests relatively low performance, with an AUC of $0.68$.
We also see that for strategy 2, Extra-Trees, RF, XGBoost, and HGBoost classifiers achieve better performance than other classifiers under most evaluation criteria.
In addition, the results confirm that Extra-Trees and HGBoost classifiers can also achieve better classification performance than RF and XGBoost in most evaluation criteria.
When training the classifier using strategy 3, we see that RF and GBoost exhibit better performance compared to other classifiers. 
GBoost and XGBoost can achieve the best AUC of $0.85$, but compared to GBoost, XGBoost shows a better recall.
The results also show that when we integrate SMOTE during training in strategies 2 and 3, we get an average recall of $0.76$ for both strategies while recall of $0.56$ for strategy 1. 
Therefore, we can make a conclusion that strategy 2 and strategy 3 would be effective predictors for screening COVID-19.

\begin{table}[!ht]
%\small
\centering
\caption{Decision matrix of the proposed method for asymptomatic category considering training strategies. Evaluation criteria into two groups based on maximization and minimization. Acc., AUC, Precision, Recall, Specificity, F1-score are expected to be the maximum; in contrast, FPR and FNR are expected to have the minimum.}
\captionsetup{justification=centering}
\resizebox{1\textwidth}{!}{
\begin{tabular}{clcccccccc} \hline
\multirow{2}{*}{\makecell[l]{Training\\strategies}} & \multirow{2}{*}{Classifiers} & \multicolumn{8}{c}{Evaluation Criteria}                                 \\ \cline{3-10}
                                     &                         & Acc.($\uparrow$) & AUC($\uparrow$)  & Precision($\uparrow$) & Recall($\uparrow$) & Specificity($\uparrow$) & F1-score($\uparrow$) & FPR($\downarrow$)  & FNR($\downarrow$)  \\ \hline \hline
\multirow{11}{*}{\rotatebox{90}{Strategy 1}}         & Extra-Trees             & 0.86 & 0.85 & 0.93      & 0.62   & 0.98        & 0.75     & 0.02 & 0.38 \\
                                     & SVM                     & 0.81 & 0.81 & 0.82      & 0.54   & 0.94        & 0.65     & 0.06 & 0.46 \\
                                     & RF                      & 0.85 & 0.81 & 0.90       & 0.62   & 0.97        & 0.73     & 0.03 & 0.38 \\
                                     & AdBoost                 & 0.82 & 0.80  & 0.82      & 0.55   & 0.94        & 0.66     & 0.06 & 0.45 \\
                                     & MLP                     & 0.81 & 0.83 & 0.84      & 0.51   & 0.95        & 0.63     & 0.05 & 0.49 \\
                                     & XGBoost                 & 0.77 & 0.68 & 0.90       & 0.33   & 0.98        & 0.48     & 0.02 & 0.67 \\
                                     & GBoost                  & 0.81 & 0.80  & 0.78      & 0.57   & 0.92        & 0.66     & 0.08 & 0.43 \\
                                     & LR                      & 0.80  & 0.78 & 0.76      & 0.57   & 0.91        & 0.65     & 0.09 & 0.43 \\
                                     & k-NN                    & 0.80  & 0.80  & 0.82      & 0.48   & 0.95        & 0.61     & 0.05 & 0.52 \\
                                     & HGBoost                 & 0.84 & 0.83 & 0.92      & 0.56   & 0.98        & 0.70      & 0.02 & 0.44 \\ \hline 
                                     %& \textbf{Weight ($W_{j}$)}                  & 0.10  & 0.06 & 0.13      & 0.06   & 0.11        & 0.07     & 0.25 & 0.22 \\ \hline \hline
\multirow{11}{*}{\rotatebox{90}{Strategy 2}}         & Extra-Trees             & 0.85 & 0.85 & 0.76      & 0.77   & 0.88        & 0.76     & 0.12 & 0.23 \\
                                     & SVM                     & 0.77 & 0.79 & 0.62      & 0.79   & 0.77        & 0.69     & 0.23 & 0.21 \\
                                     & RF                      & 0.82 & 0.84 & 0.71      & 0.77   & 0.85        & 0.74     & 0.15 & 0.23 \\
                                     & AdBoost                 & 0.78 & 0.79 & 0.64      & 0.72   & 0.81        & 0.68     & 0.19 & 0.28 \\
                                     & MLP                     & 0.80  & 0.81 & 0.69      & 0.70    & 0.85        & 0.69     & 0.15 & 0.30  \\
                                     & XGBoost                 & 0.83 & 0.84 & 0.70       & 0.79   & 0.84        & 0.75     & 0.16 & 0.21 \\
                                     & GBoost                  & 0.78 & 0.80  & 0.64      & 0.76   & 0.80         & 0.69     & 0.20  & 0.24 \\
                                     & LR                      & 0.78 & 0.79 & 0.62      & 0.78   & 0.78        & 0.69     & 0.22 & 0.22 \\
                                     & k-NN                    & 0.80  & 0.81 & 0.68      & 0.71   & 0.84        & 0.69     & 0.16 & 0.29 \\
                                     & HGBoost                 & 0.84 & 0.86 & 0.72      & 0.81   & 0.85        & 0.76     & 0.15 & 0.19 \\ \hline
                                     %& \textbf{Weight ($W_{j}$)}                  & 0.13 & 0.19 & 0.14      & 0.09   & 0.09        & 0.18     & 0.09 & 0.10  \\ \hline \hline
\multirow{11}{*}{\rotatebox{90}{Strategy 3}}         & Extra-Trees             & 0.84 & 0.83 & 0.75      & 0.74   & 0.88        & 0.74     & 0.12 & 0.26 \\
                                     & SVM                     & 0.81 & 0.83 & 0.67      & 0.79   & 0.82        & 0.72     & 0.18 & 0.21 \\
                                     & RF                      & 0.84 & 0.84 & 0.75      & 0.77   & 0.88        & 0.76     & 0.12 & 0.23 \\
                                     & AdBoost                 & 0.79 & 0.82 & 0.65      & 0.78   & 0.8         & 0.71     & 0.20  & 0.22 \\
                                     & MLP                     & 0.82 & 0.82 & 0.71      & 0.74   & 0.86        & 0.72     & 0.14 & 0.26 \\
                                     & XGBoost                 & 0.83 & 0.85 & 0.71      & 0.79   & 0.85        & 0.75     & 0.15 & 0.21 \\
                                     & GBoost                  & 0.84 & 0.85 & 0.74      & 0.76   & 0.88        & 0.75     & 0.12 & 0.24 \\
                                     & LR                      & 0.78 & 0.79 & 0.63      & 0.72   & 0.8         & 0.68     & 0.20  & 0.28 \\
                                     & k-NN                    & 0.79 & 0.79 & 0.66      & 0.70    & 0.83        & 0.68     & 0.17 & 0.30  \\
                                     & HGBoost                 & 0.83 & 0.85 & 0.71      & 0.79   & 0.85        & 0.75     & 0.15 & 0.21 \\ \hline 
                                     
                                     \hline
                                   %&  \textbf{Weight ($W_{j}$)}                & 0.10  & 0.10  & 0.09      & 0.08   & 0.12        & 0.11     & 0.19 & 0.21 \\ \hline \hline
\end{tabular}
}
\label{tab:Task 1 result}
\end{table}

\subsection{Prediction performance of symptomatic category}
\label{subsec:Prediction performance of Task 2}

Symptomatic category refers to the binary classification of symptomatic COVID-19 and non-COVID-19, where individuals are tested for COVID-19 and declare that they have a cough.
Using strategy 1, Extra-Trees and RF classifiers provide a better performance, with AUC and accuracy of $0.87$ and $0.87$, respectively.
However, the precision of the Extra-Trees classifier is better, and RF is the best in terms of recall.
In contrast, k-NN shows comparatively lower performance, with an AUC of $0.73$.
The results show that for strategy 2, the performance of Extra-Trees and MLP are almost the same.
Both classifiers provide the same AUC score, but MLP is the best at accuracy, which is $0.87$.
Furthermore, LR provides a recall value of $0.89$, which is the best among other classifiers.
For strategy 3, Extra-Trees and RF maintain almost the same performance as strategy 1, while k-NN shows the worst performance.
The results also that SMOTE can effectively deal with the class imbalance problem in the dataset, thereby improving the classification performance in strategy 2 and strategy 3.

\begin{table}[!ht]
%\small
\centering
\caption{Decision matrix of the proposed method for symptomatic category considering training strategies. Evaluation criteria into two groups based on maximization and minimization. Acc., AUC, Precision, Recall, Specificity, F1-score are expected to be the maximum; in contrast, FPR and FNR are expected to have the minimum.}
\captionsetup{justification=centering}
\resizebox{1\textwidth}{!}{
\begin{tabular}{clcccccccc} \hline
\multirow{2}{*}{\makecell[l]{Training\\strategies}} & \multirow{2}{*}{Classifiers} & \multicolumn{8}{c}{Evaluation Criteria}                                 \\ \cline{3-10}
                                     &                         & Acc.($\uparrow$) & AUC($\uparrow$)  & Precision($\uparrow$) & Recall($\uparrow$) & Specificity($\uparrow$) & F1-score($\uparrow$) & FPR($\downarrow$)  & FNR($\downarrow$)  \\ \hline \hline
\multirow{11}{*}{\rotatebox{90}{Strategy 1}}                            & Extra-Trees             & 0.87 & 0.87 & 1         & 0.8    & 1           & 0.89     & 0    & 0.20  \\
                                     & SVM                     & 0.79 & 0.78 & 0.93      & 0.72   & 0.91        & 0.81     & 0.09 & 0.28 \\
                                     & RF                      & 0.87 & 0.87 & 0.96      & 0.83   & 0.94        & 0.89     & 0.06 & 0.17 \\
                                     & AdBoost                 & 0.79 & 0.78 & 0.93      & 0.72   & 0.91        & 0.81     & 0.09 & 0.28 \\
                                     & MLP                     & 0.83 & 0.81 & 0.98      & 0.74   & 0.97        & 0.84     & 0.03 & 0.26 \\
                                     & XGBoost                 & 0.84 & 0.81 & 0.95      & 0.78   & 0.94        & 0.86     & 0.06 & 0.22 \\
                                     & GBoost                  & 0.79 & 0.75 & 0.93      & 0.72   & 0.91        & 0.81     & 0.09 & 0.28 \\
                                     & LR                      & 0.84 & 0.8  & 0.95      & 0.78   & 0.94        & 0.86     & 0.06 & 0.22 \\
                                     & k-NN                    & 0.73 & 0.75 & 0.92      & 0.63   & 0.91        & 0.75     & 0.09 & 0.37 \\
                                     & HGBoost                 & 0.77 & 0.70  & 0.87      & 0.74   & 0.81        & 0.80      & 0.19 & 0.26 \\ \hline 
                                     %& \textbf{Weight ($W_{j}$)}                  & 0.13 & 0.13 & 0.11      & 0.10    & 0.09        & 0.12     & 0.15 & 0.16 \\ \hline \hline
\multirow{11}{*}{\rotatebox{90}{Strategy 2}}                           & Extra-Trees             & 0.86 & 0.86 & 0.98      & 0.80    & 0.97        & 0.88     & 0.03 & 0.20  \\
                                     & SVM                     & 0.84 & 0.80  & 0.92      & 0.81   & 0.88        & 0.86     & 0.13 & 0.19 \\
                                     & RF                      & 0.84 & 0.85 & 0.95      & 0.78   & 0.94        & 0.86     & 0.06 & 0.22 \\
                                     & AdBoost                 & 0.80  & 0.79 & 0.91      & 0.76   & 0.88        & 0.83     & 0.13 & 0.24 \\
                                     & MLP                     & 0.87 & 0.86 & 0.94      & 0.85   & 0.91        & 0.89     & 0.09 & 0.15 \\
                                     & XGBoost                 & 0.81 & 0.84 & 1         & 0.70    & 1           & 0.83     & 0    & 0.30  \\
                                     & GBoost                  & 0.86 & 0.83 & 0.94      & 0.83   & 0.91        & 0.88     & 0.09 & 0.17 \\
                                     & LR                      & 0.86 & 0.83 & 0.89      & 0.89   & 0.81        & 0.89     & 0.19 & 0.11 \\
                                     & k-NN                    & 0.72 & 0.74 & 0.97      & 0.57   & 0.97        & 0.72     & 0.03 & 0.43 \\
                                     & HGBoost                 & 0.77 & 0.74 & 0.93      & 0.69   & 0.91        & 0.79     & 0.09 & 0.31 \\ \hline 
                                     %& \textbf{Weight ($W_{j}$)}                  & 0.09 & 0.16 & 0.14      & 0.09   & 0.10         & 0.08     & 0.15 & 0.18 \\ \hline \hline
\multirow{11}{*}{\rotatebox{90}{Strategy 3}}                           & Extra-Trees             & 0.84 & 0.83 & 1         & 0.74   & 1           & 0.85     & 0    & 0.26 \\
                                     & SVM                     & 0.80  & 0.79 & 0.93      & 0.74   & 0.91        & 0.82     & 0.09 & 0.26 \\
                                     & RF                      & 0.87 & 0.85 & 0.96      & 0.83   & 0.94        & 0.89     & 0.06 & 0.17 \\
                                     & AdBoost                 & 0.83 & 0.80  & 0.95      & 0.76   & 0.94        & 0.85     & 0.06 & 0.24 \\
                                     & MLP                     & 0.83 & 0.78 & 0.88      & 0.83   & 0.81        & 0.86     & 0.19 & 0.17 \\
                                     & XGBoost                 & 0.84 & 0.81 & 0.92      & 0.81   & 0.88        & 0.86     & 0.13 & 0.19 \\
                                     & GBoost                  & 0.88 & 0.87 & 0.98      & 0.83   & 0.97        & 0.90      & 0.03 & 0.17 \\
                                     & LR                      & 0.78 & 0.76 & 0.95      & 0.69   & 0.94        & 0.80      & 0.06 & 0.31 \\
                                     & k-NN                    & 0.69 & 0.71 & 0.97      & 0.52   & 0.97        & 0.67     & 0.03 & 0.48 \\
                                     & HGBoost                 & 0.83 & 0.80  & 0.91      & 0.80    & 0.88        & 0.85     & 0.13 & 0.20  \\ \hline 
                                     
                                     \hline
                                     %& \textbf{Weight ($W_{j}$)}                  & 0.07 & 0.09 & 0.10       & 0.07   & 0.09        & 0.07     & 0.15 & 0.37 \\ \hline \hline
\end{tabular}
}
\label{tab:Task 2 result}
\end{table}

\begin{table}[!ht]
%\small
\centering
\caption{Evaluation criteria and weights based on the entropy of all categories.}
\captionsetup{justification=centering}
\resizebox{1\textwidth}{!}{
\begin{tabular}{lccccccccc} \hline
\multirow{2}{*}{Category} & \multirow{2}{*}{\makecell[l]{Training\\strategies}} & \multicolumn{8}{c}{Evaluation criteria}                                 \\ \cline{3-10}
                                     &                         & Acc. & AUC  & Precision & Recall & Specificity & F1-score & FPR  & FNR  \\ \hline \hline
\multirow{3}{*}{Asymptomatic}         & Strategy 1                  & 0.10  & 0.06 & 0.13      & 0.06   & 0.11        & 0.07     & 0.25 & 0.22 \\ %\cline{2-10} 
 & Strategy 2                  & 0.13 & 0.19 & 0.14      & 0.09   & 0.09        & 0.18     & 0.09 & 0.10  \\ %\cline{2-10} 
 &  Strategy 3                & 0.10  & 0.10  & 0.09      & 0.08   & 0.12        & 0.11     & 0.19 & 0.21 \\ \hline
 
\multirow{3}{*}{Symptomatic}         & Strategy 1                  & 0.13 & 0.13 & 0.11      & 0.10    & 0.09        & 0.12     & 0.15 & 0.16 \\ %\cline{2-10} 
 & Strategy 2                  & 0.09 & 0.16 & 0.14      & 0.09   & 0.10         & 0.08     & 0.15 & 0.18  \\ %\cline{2-10} 
 &  Strategy 3                & 0.07 & 0.09 & 0.10       & 0.07   & 0.09        & 0.07     & 0.15 & 0.37 \\ \hline 
 
 \hline
 
\end{tabular}
}
\label{tab:Task 1 and Task 2 entropy weight}
\end{table}

\subsection{Model selection using ensemble-based MCDM}
\label{subsec:Model selection using ensemble-based MCDM}

This section presents the results of selecting an optimal diagnostic model for COVID-19 through ensemble-based MCDM.
Table~\ref{tab:Task 1 result} and Table~\ref{tab:Task 2 result} provide decision matrices considering all training strategies of asymptomatic and symptomatic categories, respectively. 
Table~\ref{tab:Task 1 and Task 2 entropy weight} shows the entropy-based weights of the decision matrix based on all evaluation criteria (Algorithm~\ref{entropy} shows the steps required for calculation).
FPR and FNR (in asymptomatic and symptomatic) maintain the maximum weight of strategy 1 and strategy 3, while AUC maintains the maximum weight of strategy 2 in the two tasks.
According to the results, the criterion with the highest weight is the most important criterion, and the least important criterion has a lower weight value.
Next, multiply the normalized decision matrix and the weight to obtain the weighted normalized decision matrix, as described in step 2 in Algorithm~\ref{TOPSIS}.
Furthermore, Table~\ref{tab:Idea Best and Ideal Worst} shows the results of the ideal best value and the ideal worst value generated from the weighted decision matrix, as shown in step 3 of Algorithm~\ref{TOPSIS} and Equations~\ref{eqn:ideal best} and~\ref{eqn:ideal worst}. 

According to Table~\ref{tab:Idea Best and Ideal Worst}, each COVID-19 diagnostic model shows the difference of each criterion in respect of the ideal best and worst values.
Before calculating the relative closeness value, we need to measure two separations, $S^{+}$ and $S^{-}$, which reflect how close each classifier is to the ideal best and worst (see step 4 of the Algorithm~\ref{TOPSIS}).
The hypothesis that influences the selection of the best model is that the best model's $S^{+}$ value is the minimum compared to the other model's $S^{+}$ value. In contrast, the best model's $S^{-}$ value is relatively higher compare to other model's $S^{-}$ value.

Table~\ref{tab:Ensemble MCDM result} shows the relative closeness value ($C_{mj}$) of each training strategy of the ten classifiers using step 5 of Algorithm~\ref{TOPSIS}.
We further integrate these relative closeness values into ensemble methods, such as soft ensemble and hard ensemble, to rank models.
In the case of the soft ensemble, we take the average of the relative closeness values, and give the final ranking based on the average; the highest average value reflects the best model.
In this way, we have seen Extra-trees become the top model for asymptomatic and symptomatic categories.
On the other hand, for hard ensemble, we assign points ($C_{mj}P$) to each $C_{mj}$ value mapped from $1$ to $10$, where the highest point is assigned to the highest $C_{mj}$.
However, if two or more models have the same $C_{mj}$ value, we assign the same point.
After summing up all the points, we got the top-ranked model.
It can be seen from Table~\ref{tab:Ensemble MCDM result} that the results of the hard ensemble reflect that HGBoost is the best for asymptomatic, and for symptomatic, the Extra-Trees classifier is at the top. 

After analyzing the results of integrating MCDM (Table~\ref{tab:Ensemble MCDM result}), we can say that the proposed method using Extra-trees and HGBoost classifiers is better than other classifiers.
Table~\ref{tab:Comparision result} shows the comparison of the detection of asymptomatic and symptomatic COVID-19 from cough samples using Extra-Trees and HGBoost classifiers based on our proposed method.
For the asymptomatic category, we see that our proposed method's AUC using HGBoost classifier is higher than Extra-Trees classifier.
Extra-Trees classifier shows higher precision, but the AUC and recall rate lag behind HGBoost.
When comparing the precision results, for the symptomatic category, we see that the Extra-Trees classifier shows impressive results when classifying COVID-19 symptomatic cough, with a precision rate of $1$.
On the other hand, HGBoost achieves a recall of $0.80$, which is higher than Extra-trees.

Figure~\ref{fig:confusion matrix} shows the confusion matrices of the proposed method considering Extra-Trees classifier for all training strategies.
In Figure~\ref{fig:confusion matrix} (b)-(c), for COVID-19 asymptomatic cough detection, strategy $2$ provides results that are $3\%$ better than strategy $3$.
Moreover, the proposed method can effectively detect non-COVID-19 asymptomatic coughs; whether in strategy $2$ or strategy $3$, it can provide identical performance.
Although strategy $1$ shows relatively low performance compared to other strategies for asymptomatic COVID-19 cough detection, strategy $1$ outperforms other strategies in a case of symptomatic category.   
When comparing strategy $2$ and strategy $3$ for non-COVID-19 symptomatic cough detection, Extra-Trees classifier provides excellent results through strategy 3.
In addition, for asymptomatic and symptomatic COVID-19 cough detection, training strategy $2$ outperforms strategy $3$, ranging from $3\%$ to $6\%$.

\begin{table}[!ht]
%\small
\normalsize
\centering
\caption{The results of the ideal best and the ideal worst value of each task for each training strategy.}
\captionsetup{justification=centering}
%\resizebox{1\textwidth}{!}{
\begin{tabular}{llcccccc} \hline
\multirow{2}{*}{Category}       & \multirow{2}{*}{\makecell[l]{Evaluation\\criteria}} & \multicolumn{2}{c}{Strategy 1} & \multicolumn{2}{c}{Strategy 2} & \multicolumn{2}{c}{Strategy 3} \\ \cline{3-8}
                        &           & $V^{+}$        & $V^{-}$        & $V^{+}$        & $V^{-}$    & $V^{+}$        & $V^{-}$        \\ \hline\hline
\multirow{8}{*}{Asymptomatic} & Acc.                    & 0.032         & 0.028          & 0.042         & 0.038          & 0.031         & 0.029          \\
                        & AUC                     & 0.020          & 0.016          & 0.064         & 0.059          & 0.034         & 0.032          \\
                        & Precision               & 0.045         & 0.037          & 0.049         & 0.040           & 0.032         & 0.027          \\
                        & Recall                  & 0.023         & 0.013          & 0.030          & 0.026          & 0.025         & 0.022          \\
                        & Specificity             & 0.037         & 0.034          & 0.029         & 0.025          & 0.041         & 0.037          \\
                        & F1-score                & 0.025         & 0.016          & 0.059         & 0.053          & 0.036         & 0.032          \\
\hline 
                        
                        \hline         
\end{tabular}
%}
\label{tab:Idea Best and Ideal Worst}
\end{table}

\begin{table}[!ht]
%\small
\centering
\caption{Results of MCDM with integration of ensemble.}
\captionsetup{justification=centering}
\resizebox{1\textwidth}{!}{
\begin{tabular}{clccc|cc|ccccc} \hline
Category   & Classifiers & \multicolumn{3}{c|}{Relative Closeness Scores} & \multicolumn{7}{c}{Ensemble}              \\ \hline
                        &             &               &               &               & \multicolumn{2}{c|}{Soft} & \multicolumn{5}{c}{Hard}                     \\ \cline{6-12}
                        &             & $C_{m1}$           & $C_{m2}$           & $C_{m3}$           & Avg.($C_{mj}$)      & Rank     & $C_{m1}P$ & $C_{m2}P$ & $C_{m3}P$ & Total($C_{mj}P$) & Rank \\ \hline \hline
\multirow{10}{*}{Asymptomatic} & Extra-Trees & 1                         & 0.806 & 0.701 & 0.835     & \textbf{\underline{1}}           & 10     & 10     & 6      & 26           & 2         \\
                         & SVM         & 0.478                     & 0.370  & 0.535 & 0.461      & 7           & 3      & 4      & 4      & 11           & 8         \\
                         & RF          & 0.871                     & 0.683 & 0.867 & 0.807      & 3           & 8      & 7      & 10     & 25           & 3         \\
                         & AdBoost     & 0.483                     & 0.256 & 0.422 & 0.387      & 9           & 4      & 1      & 3      & 8            & 9         \\
                         & MLP         & 0.579                     & 0.428 & 0.614 & 0.540     & 5           & 6      & 6      & 5      & 17           & 5         \\
\end{tabular}
    }
\label{tab:Ensemble MCDM result}
\end{table}

\begin{table}[!ht]
\normalsize
%\small
\centering
\caption{Comparison of the proposed methods for COVID-19 cough detection.}
\captionsetup{justification=centering}
%\resizebox{10cm}{!}{
%
%}
\caption{Normalized confusion matrices of Extra-Tree classifiers with 10-fold cross-validation for all training strategies. Figures (a)-(c) represent the confusion matrix of asymptomatic categories, and for symptomatic categories, the confusion matrices are (d)-(f). The sum of each class is equal to $1$. Note that $0$ represents COVID-19 and $1$ represents Non-COVID-19 cough.
}
\label{fig:confusion matrix}

\end{figure}

\subsection{Feature dimension reduction}
\label{Feature dimension reduction}

We analyze the effect of feature dimensionality reduction on asymptomatic and symptomatic categories.
In this regard, we use the cross-validated recursive feature elimination (RFECV). It is based on the feature importance weights and cross-validation to automatically adjust the number of selected features.
We use three supervised learning estimators, i.e., Extra-Trees, LinearSVC, and LDA, while fitting the method that provides information about feature importance.
Figure~\ref{fig:Cambridge Feature Reduction-a} shows the optimal number of feature selections using different estimators for the asymptomatic category.
Extra-Trees estimator achieves a fairly good AUC score, exceeding $0.80$ while maintaining the best features. However, other estimators such as LinearSVC and LDA achieve lower AUC than Extra-Trees.
In this regard, the total number of best features generated using Extra-Trees estimator is $38$, but the total number of best features generated using LinearSVC and LDA estimators are $6$ and $78$, respectively. 

For symptomatic, we observe a similar trend in Figure~\ref{fig:Cambridge Feature Reduction-b}. Extra-Trees obtains a higher AUC than LinearSVC and LDA while retaining the best features.
Extra-Trees estimator selects a total of $6$ best features, while LinearSVC and LDA estimators select a total of $1$ and $3$ best features, respectively.
Here, we observe that both categories (i.e., asymptomatic and symptomatic) produce comparable AUC scores while using Extra-Trees as an estimator, but the symptomatic category retains fewer features than the asymptomatic category.

\begin{figure}[!ht]
  \centering
  \captionsetup[subfigure]{font=scriptsize,labelfont=scriptsize}
  \resizebox{1\textwidth}{!}{
  \begin{subfigure}[b]{0.35\linewidth}
    \includegraphics[width=\textwidth]{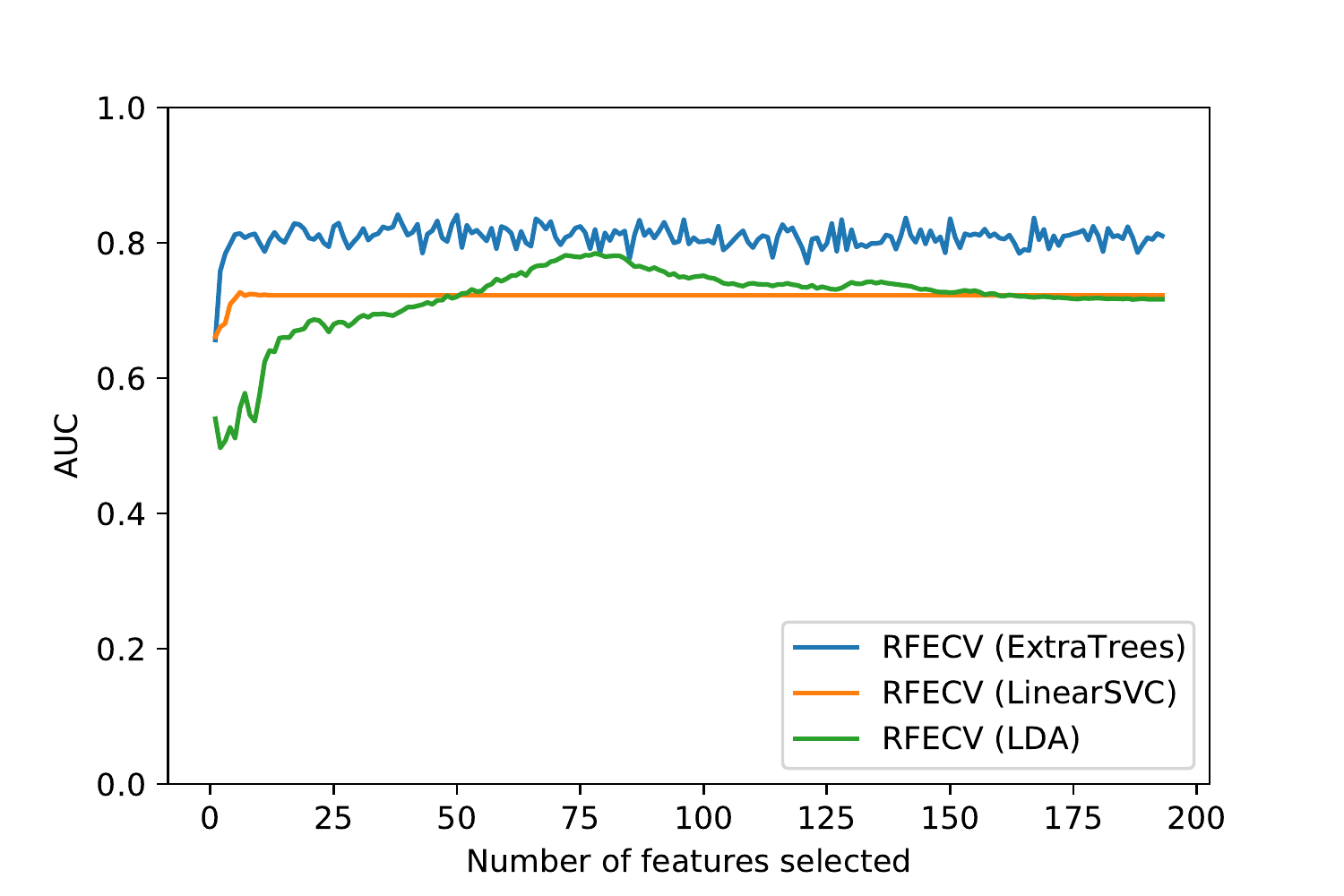}
    \captionsetup{justification=centering}
    \caption{RFECV for the Cambridge asymptomatic category}
    \label{fig:Cambridge Feature Reduction-a}
  \end{subfigure}
  \begin{subfigure}[b]{0.35\linewidth}
    \includegraphics[width=\textwidth]{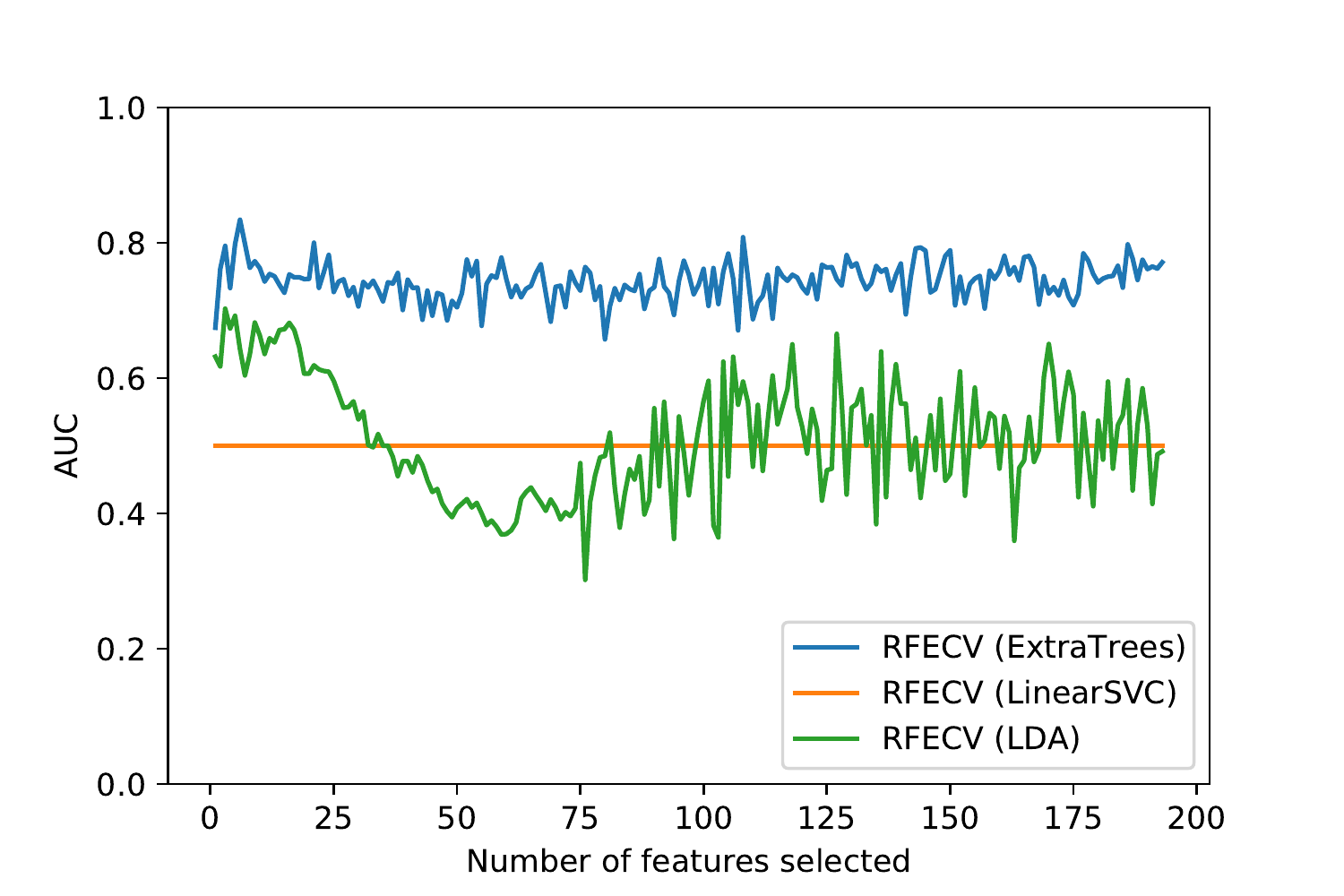}
    \captionsetup{justification=centering}
    \caption{RFECV for the Cambridge symptomatic category}
    \label{fig:Cambridge Feature Reduction-b}
  \end{subfigure}
      }
  %\captionsetup{justification=centering}
  \caption{Optimal numbers of feature selection using recursive feature elimination with cross-validation for Cambridge asymptomatic and symptomatic categories. Note that RFECV stands for Recursive Feature Elimination with Cross-Validation.}
     
  \label{fig:Cambridge Feature Reduction}
\end{figure}

\subsection{Comparison}
\label{Comparison between the proposed model and the other model}

Table~\ref{tab:Comparision result with feature reduction} shows the comparison between our proposed model with integrating feature selection and the state-of-the-art models for detecting COVID-19 from cough samples.
The purpose is not to do a direct comparison except the work~\cite{1}, because the implementation details of other works are not available or the dataset is different from us.
%which means the number of cough samples is also different.
When comparing "with feature selection" with "no feature selection" approach for the asymptomatic category, we see that the AUC and recall value of our proposed Extra-Trees classifier with feature selection score higher, $0.88$ and $0.81$, respectively.
On the other hand, for symptomatic category, our proposed method with feature selection provides significantly better results than no feature selection.
Note that, results of "no feature selection" are reported in Table~\ref{tab:Comparision result} while Table~\ref{tab:Comparision result with feature reduction} shows the results of "with feature selection". 
Obviously, when considering the feature selection step, the performance of the Extra-Trees classifier is shown relatively better than that of the HGBoost classifier.
When comparing with Brown et al.~\cite{1} in the asymptomatic category, we see that our proposed method's AUC and recall using Extra-Trees classifier is higher than that.
What's more, HGBoost achieves a precision of $0.76$, which is higher than others.
HGBoost shows better result than the previous study~\cite{1}, but the AUC and recall rate lag behind Extra-Trees.

As we have observed from empirical evaluation, for the symptomatic category, the proposed method using Extra-Trees classifier outperforms the previous study~\cite{1}.
We also see that the Extra-Trees classifier shows impressive results when classifying COVID-19 symptomatic cough, with a precision rate of $1$.
On the other hand, Brown et al.~\cite{1} achieved a recall of $0.90$, which is comparable to Extra-Trees.
In addition, the overall precision of the model~\cite{38} is $0.87$, and the precision of the proposed method to symptom category reporting is $1$.
However, the dataset setting of the symptomatic category is different from ours.

For Coswara dataset, the precision and recall of the Extra-Trees classifier are $0.70$ and $0.58$, respectively.
The HGBoost classifier shows better AUC and precision than the Extra-Trees classifier, but it lags significantly behind when comparing recall rates.
For Virufy dataset, the AUC, precision, and recall rate for detecting COVID-19 are $0.94$, $0.89$, and $0.98$, respectively, which indicates that our proposed model has high detection performance when considering the HGBoost classifier.
In the case of integrating Virufy with the NoCoCoDa dataset, our proposed model achieves higher AUC values of $0.97$ and $0.98$ for Extra-Trees and HGBoost respectively, which means that our model has a lower false negative and false positive rate.
In addition, the recall rate of the HGBoost classifier is as high as $0.98$.
Such a high recall rate ensures that our proposed model will have a very low false negative result for COVID-19, making it a suitable screen for detecting COVID-19.
The detection performance between us and Melek~\cite{67} is almost the same, but Melek~\cite{67} considered $59$ COVID-19 samples from the NoCoCoDa dataset, while we considered all $73$ COVID-19 samples.
%Other studies~\cite{37,18} used their own dataset, which is not publicly available.
%The precision and recall rates of coughs for COVID-19 and non-COVID-19 patients are $0.90$ in~\cite{37}, while in~\cite{18} precision and recall rates are $0.91$ and $0.95$, respectively. 

\begin{table}[!ht]
\normalsize
%\small
\centering
\caption{Comparison of our proposed approach with the state-of-the-art approaches}
\captionsetup{justification=centering}
%\resizebox{1\textwidth}{!}{
\begin{tabular}{cclccc} \hline
 \multicolumn{2}{c}{Dataset} & Method   & AUC & Precision & Recall \\ \hline \hline
\multirow{6}{*}{Cambridge} & \multirow{3}{*}{Asymptomatic} & Brown et al.~\cite{1}    & 0.80   & 0.72         & 0.69      \\ %\cline{2-5}
                       & & \makecell[l]{Proposed (RFECV + Extra-Trees) } & \textbf{0.88}   & 0.75         & \textbf{0.81}      \\ %\cline{2-5}
                    &    & \makecell[l]{Proposed (RFECV + HGBoost) }  & 0.85   & \textbf{0.76}      & 0.73  \\ \cline{2-6}
&\multirow{3}{*}{Symptomatic} & Brown et al.~\cite{1}    & 0.87   & 0.70         & 0.90      \\ %\cline{2-5}
                    &    & \makecell[l]{Muhammad et al.~\cite{38} } & -   & 0.87         & 0.82      \\ %\cline{2-5}
                    &    & \makecell[l]{Proposed (RFECV + Extra-Trees) } & \textbf{0.95}   & \textbf{1}         & \textbf{0.91}      \\ %\cline{2-5}
                      &    & \makecell[l]{Proposed (RFECV + HGBoost) }  & 0.81   & 0.93      & 0.80  \\
                        \hline
                        
\multicolumn{2}{c}{ \multirow{2}{*}{Coswara}} & \makecell[l]{Proposed (RFECV + Extra-Trees) } & 0.64   & 0.70   & 0.58    \\ %\cline{2-5}
                    %& & \makecell[l]{Proposed (RFECV + Extra-Trees) } & 0.64   & 0.70   & 0.58    \\ %\cline{2-5}
                      &   & \makecell[l]{Proposed (RFECV + HGBoost)} & 0.66   & 0.76         & 0.47      \\ %\cline{2-5}
                     % &   & \makecell[l]{Madhurananda et al.~\cite{41}} & 0.98   & -         & 0.93      \\ %\cline{2-5}
                        \hline

\multicolumn{2}{c}{ \multirow{2}{*}{Virufy}} & \makecell[l]{Proposed (RFECV + Extra-Trees) } & 0.92   & \textbf{0.89}   & 0.88    \\ %\cline{2-5}
                      &   & \makecell[l]{Proposed (RFECV + HGBoost)} & \textbf{0.94}   & \textbf{0.89}         & \textbf{0.98}      \\ %\cline{2-5}
                        \hline
\multicolumn{2}{c}{ \multirow{2}{*}{Virufy+NoCoCoDa}} & \makecell[l]{Melek~\cite{67}} & \textbf{0.99}   & 0.99         & 0.97      \\ %\cline{2-5}
                   &    & \makecell[l]{Proposed (RFECV + Extra-Trees) } & 0.97   & \textbf{1}   & 0.92    \\ %\cline{2-5}
                      &   & \makecell[l]{Proposed (RFECV + HGBoost)} & 0.98   & 0.99         & \textbf{0.98}      \\ %\cline{2-5}
                      
                       % \hline
%\multicolumn{2}{c}{ \multirow{2}{*}{Other Datasets}} & \makecell[l]{Ankit et al.~\cite{37}} & -   & 0.90         & 0.90      \\ %\cline{2-5}
    %               &    & \makecell[l]{Imran et al.~\cite{18} } & -   & 0.91   & 0.95    \\ %\cline{2-5}
                        \hline
                        
                        \hline  
\multicolumn{6}{l}{-\textbf{Bold} values indicate that they are the highest.}\\
\end{tabular}
%}
\label{tab:Comparision result with feature reduction}
\end{table}

\vspace{10mm}

\section{Conclusion and Future Work}
\label{sec:conclusion}

In this paper, we present an ensemble based MCDM method for detecting COVID-19 from cough samples. In particular, we address the challenge to select the best classification model considering eight evaluation criteria where there exist a variation among these evaluation criteria.
At first, we generate features that stem from the audio analysis of cough samples.
In the training process, we consider three training strategies with different parameter settings to assess the effectiveness of various aspects of the proposed method.
After that, we construct a decision matrix of ten ML-driven classifiers with eight evaluation criteria for each training strategy.
Next, the proposed method integrates TOPSIS to rank the models of each training strategy, where the weight of the evaluation criteria is calculated using entropy. 
Subsequently, using ensemble methods, namely soft ensemble and hard ensemble, the best COVID-19 diagnostic model is identified based on the quantitative information of the measurement standards (such as average and counting votes corresponds to relative closeness value).
The reason behind choosing the ensemble strategy is that it reduces the bias in selecting the best model as the relative closeness values of different training strategies greatly affect the ranking of the model.
%Another reason is that it is a challenge to select the best model considering several evaluation criteria since there is the issue of variation among these evaluation criteria.  
Our empirical evaluation shows that the proposed method considering Extra-Trees and HGBoost classifiers provide better result. 
It also confirmed that the tree-based ensemble learning classifiers performed better than the non-tree-based ensemble learning classifiers. 
Therefore, we believe our findings could contribute usefully in detecting COVID-19 infection.
%and the extensive acceptance of AI-driven automated systems in medical applications.

In future work, we will study cross-institutional dataset using more COVID-19 cough cases to make our proposed method more robust.
In addition, the tracking of the progress of COVID-19 and the analysis of the severity of COVID-19 cough behavior can be explored to study further improvements in performance. 
%We will spring up a mobile application to prognosticate whether the disease will become deadly through analyzing a patient’s short-term medical record if the patient exposes any clinical symptoms related to COVID-19 coughing. 
%Therefore, this may be an alternative way to halt the second wave of the COVID-19 pandemic.

\section*{Declaration of Competing Interest}
\label{Declaration of Competing Interest}

All authors declare that there is no conflict of interest in this work

\section*{Acknowledgement}
\label{sec:Acknowledgement}

The authors are grateful to the University of Cambridge for providing their COVID-19 sound dataset.
We would like to thank Madison Cohen-McFarlane of Carleton University for sharing the NoCoCoDa dataset. Cambridge University and Carleton University assume no responsibility for the results reported in this article.

\bibliographystyle{unsrt}
\bibliography{reference}

\end{document}